# Institutional Books — Visual Elements: An open-source pipeline for extracting, classifying, deduplicating, and captioning visual elements from digital book collections.

**Jimmy Mendez** [a], **Matteo Cargnelutti** [a], **David Lowry-Duda** [a], **Catherine Brobston** [a], **Salwa Ismail** [b], **Greg Leppert** ✉ [a], **Amanda Watson** [c], **Jonathan Zittrain** [d]

[a] Institutional Data Initiative, Harvard Law School Library
[b] Harvard Library
[c] Harvard Law School Library
[d] Harvard Law School, Harvard School of Engineering and Applied Sciences, Harvard Kennedy School

## Abstract

Historical book collections contain rich visual elements—such as illustrations, photographs, engravings, and decorative art—that are frequently under-explored in large-scale digitization projects. While Optical Character Recognition (OCR) has standardized the extraction of textual content, these visual components offer a layer of nuance and context that remains largely untapped by automated text extraction workflows. This technical report introduces **Institutional Books — Visual Elements**, an open-source **end-to-end pipeline** for detecting, classifying, deduplicating, and captioning visual elements from historical book collections. Alongside this pipeline, we release **an initial dataset of 22.6 million visual elements** extracted from the 983,004 scanned volumes that comprise the **Institutional Books: Harvard Library dataset**. This work contributes to ongoing, community-wide efforts to enable new use cases for digitized library collections through computational access, from artificial intelligence model training to digital humanities research.

✉ Corresponding author: gleppert@law.harvard.edu

# 1 Introduction

The work we introduce with this technical report builds on our initial release of the Institutional Books: Harvard Library dataset (Cargnelutti *et al.*, 2025), which made close to one million books from Harvard Library's Google Books collection available for computational access. As AI systems bring computational search and research within reach for broad swaths of society, we posit that computational access to institutional collections is an increasingly powerful gateway to new use cases and renewed relevance for those materials. Where our initial release focused on text and metadata, this additional pipeline and dataset aims to broaden the spectrum of possible computational uses for this type of collection by detecting, extracting, classifying, deduplicating, and captioning visual elements from across book corpora.

Data diversity, scale and quality directly impact model performance, task adaptability, and genericity (Gao *et al.*, 2020; Gunasekar *et al.*, 2023; Hoffmann *et al.*, 2022; Kaplan *et al.*, 2020; Longpre et al., 2024). By nature of the data they are commonly trained on, Vision-Language Models (VLMs) tend to reflect the predominantly contemporary visual culture of the internet (Impett and Offert, 2022; Schneider, 2026). Recent work has shown that they can struggle to process historical content (Strafforello *et al.*, 2025), exhibit degraded performance under temporal distribution shift (Pégeot *et al.*, 2025), and are prone to cultural anachronism and inaccurate temporal reasoning (Barancová, Wevers and Noord, 2023; Ranjan *et al.*, 2026; Tekaya, Waldner and Zeppelzauer, 2025). We hypothesize that making visual elements with rich metadata from historical book collections available for computational access can help bridge known gaps in the way models "see" and "understand" the world, paving the way for these models to better assist libraries in working with historical materials. Our hope is that this results in a virtuous cycle in which greater data availability produces stronger model capabilities for working with these materials, which in turn unlock more data.

While extracting visual elements out of the approximately 400 million page scans that comprise Harvard Library's Google Books collection is necessarily computationally intensive, we approached this project with frugal computing principles in mind (Vanderbauwhede, 2023). For each method and model we developed as part of this pipeline, we started from the smallest viable intervention and only escalated computational expense when strictly necessary to produce high quality data. We posit that this relative restraint helps with the reproducibility of our work, and it is our hope that it will allow more knowledge institutions to perform this type of advanced data extraction at scale. Section 9 of this technical report provides details on the results of that design process.

The processing steps we developed as part of this pipeline range from well-established methods to experimental practices. AI-assisted image captioning (Section 6) is an example of the latter. We expect the outputs of this captioning experiment to help users navigate the dataset, but also enable downstream research on how VLMs process visual elements from historical collections. Clear documentation and careful data labeling is crucial in that context. Section 8 of this report describes how our pipeline's output clearly distinguishes data points that were directly ported from source materials from those that were generated or are experimental.

With this initial release of our pipeline, models, and resulting dataset, we seek to contribute to an iterative, community-based practice of surfacing overlooked sources of meaningful visual elements — to the benefit of the library, AI, and research communities alike.

# 2 Contributions

With this technical report, we introduce this initial set of contributions:

1. **An open-source pipeline** for detecting, classifying, deduplicating, and captioning visual elements from historical book collections page scans.
2. **A dataset** containing 22,622,060 visual elements extracted from the 983,004 volumes in the Institutional Books: Harvard Library dataset. Each element is accompanied by a classification, an image embedding, and an experimental, LLM-generated caption when applicable. The dataset also comes with a SKILL.md file to facilitate agentic use. The aggregate captioning data consists of 766,992,447 `o200k_base` text tokens.
3. **A set of fine-tuned models** for visual element detection, classification, and orientation correction built on top of open-source vision architectures.
4. **An analysis** of the characteristics of the dataset, including a comparison between the properties of its visual data and the text-only data from the Institutional Books: Harvard Library dataset.

The dataset and models are available on Hugging Face:
https://huggingface.co/collections/institutional/institutional-books.

The pipeline's source code is available on GitHub:
https://github.com/institutional/institutional-books-visual-elements-pipeline.

# 3 Detection

## 3.1 Methodology

The detection step serves as the foundational component of the visual elements extraction pipeline, responsible for localizing regions of interest within document scans. Detection outputs directly determine the scope of all downstream analysis. Indeed, classification, embedding generation, captioning, and deduplication all operate exclusively on previously-detected regions. Detecting visual elements at the scale of the Institutional Books collection requires using a method that can achieve both high accuracy and high throughput for this specialized task. After initial review of the available data for the collection, we determined that a machine-learning-based model would be required to extract the desired crops. While a vision-language model may be able to achieve high accuracy, running inference using a transformer-based architecture against the entire collection would represent a significant computational expense, one that would be out of reach for most knowledge institutions. Moreover, the main benefits of VLMs, such as few-shot learning and robustness on datasets with weak inductive biases (Danish *et al.*, 2026), did not apply to our dataset, as many of the historical pages follow consistent visual and typographic conventions. Because the task of detecting and classifying images from a digitized books collection is domain-specific and exhibits strong inductive biases, we chose a CNN-based approach.

### 3.1.1 Training

Establishing a training set initially involved manually annotating 109 randomly-selected samples from the collection in order to fine-tune an object detection CNN before scaling the training set up to 4,528 samples through partial automation. After training a model on this initial training set, we trained intermediary models, both as a way to adjust our training strategy (adjusting hyperparameters and data augmentation settings) and to pre-annotate additional batches of samples, which we reviewed and corrected manually. We used the VGG Image Annotator (VIA3) (Dutta and Zisserman, 2019) to edit and correct all of our annotations, regardless of their origin (fully manual versus pre-annotated using an intermediary model). The iterative manual review and annotation process took several weeks before landing on a final training set.

The final training set consisted of 4,528 page images sourced from 271 distinct volumes in the collection. The dataset contained 6,103 bounding box annotations of type-agnostic visual elements. The training set also contained negative examples: images where no visual elements were present. These negative examples were included to ensure the model correctly learned to suppress spurious detections in pages without any visual elements and to calibrate its decision boundary between true positives and background. This was especially important as most of the pages in the collection did not contain any visual elements.

The dataset was randomly shuffled and partitioned into training (80%), validation (10%), and test (10%) splits. These splits were done at the page level, where the pages were shuffled and assigned a split randomly. For the visual element extraction task, annotations were merged into a single object class to formulate the problem as a binary detection task rather than multi-class classification. By training a single-class detector rather than a multi-class model, we decouple localization from categorization, allowing each task to be optimized independently. This architectural decision enabled the detection model to focus on identifying visual elements broadly, while delegating fine-grained classification to a separate downstream model. This also allowed us to account for natural class imbalance independently.

Note that because the collection contains likely duplicates (Section 5), both within volumes and across the collection, there is a residual risk of indirect contamination.

After iterating over different object detection models, we chose a YOLO model variant—YOLO26n with 2.4 million parameters—for its interoperability, limited footprint, and ability to adjust to a domain-specific task with limited training data (Jocher *et a*l. 2026; Sapkota *et al.*, 2026). The model was trained on preprocessed images of 800x800 pixels, with a batch size of 30 scan images per GPU for up to 400 epochs and a patience of 50. The model was trained with an elevated box loss weight of 8.3 and a DFL loss of 1.5 to prioritize precise localization over classification, reflecting the one-class detection approach. The scale augmentation factor of 0.7 introduced moderate size variations during training to improve robustness across document images of varying resolutions. These settings were chosen by iteratively testing and evaluating the model produced.

Training utilized up to four NVIDIA L40S GPUs in parallel with automatic device selection. Model checkpoints were saved based on validation performance, with the best-performing weights retained for inference, which was at epoch 200. The total training time was 1.4 hours. Table 1 shows the metrics achieved on the test set.

*Table 1: Detection model performance*

| Metric | Value |
|---|---|
| Total Pages (test) | 453 |
| Annotations | 658 |
| Precision | 0.95 |
| Recall | 0.96 |

### 3.1.2 Inference

The image size was set to 640x640 pixels for inference. This was chosen to increase throughput and no meaningful accuracy degradation was observed. The confidence threshold was set to 0.3, a deliberately conservative choice intended to maximize recall, even at the risk of introducing more false positives, which were later pruned using classification confidence scores. The relatively low NMS IoU threshold of 0.2 was used to retain distinct yet spatially close visual elements, which are typical of densely illustrated historical documents as shown in Figure 3. These values were chosen based on iterative evaluation, and focused on the most common case in this collection, which are sparse documents.

Figures 1 and 2 show a pair of heatmaps generated using EigenCAM (Muhammad and Yeasin, 2020; rigvedrs, 2026) against our detection model. The heatmaps represent the first principal component of the 2D activations in a neural network, without taking class discrimination into account.

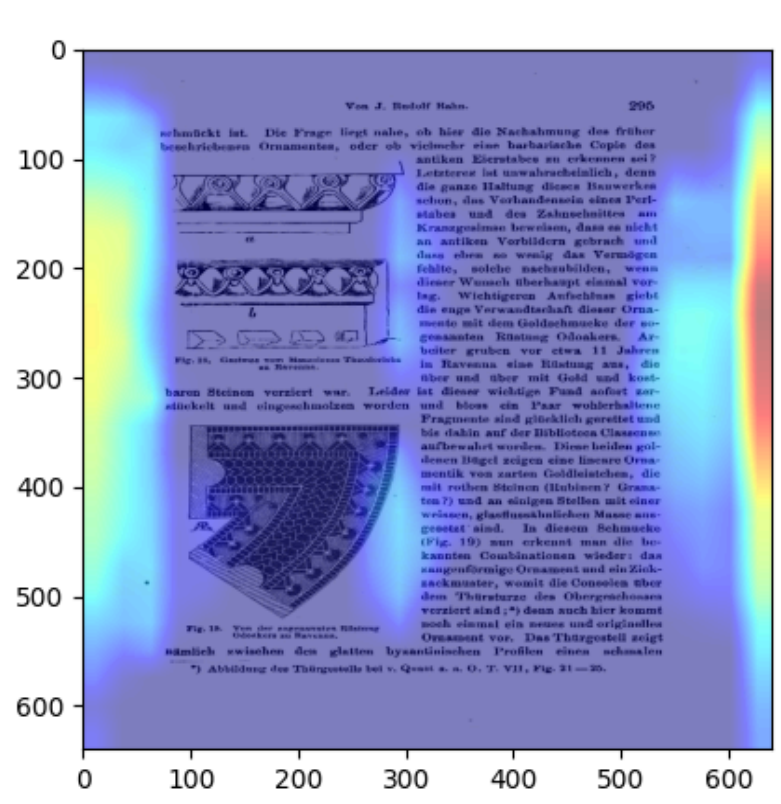

*Figure 1:* Heatmap
The detection model seems to pay attention to the whitespace around the page to determine if there is a visual element. Here, it correctly identifies 3 distinct objects.

*Figure 2:* Heatmap
The detection model seems to pay attention to the whitespace around the page to determine if there is a visual element. Here, it correctly identifies 2 distinct objects.

### 3.1.3 Image preprocessing

Page scans presented as either TIFF or JP2 files and were decoded as numpy arrays prior to inference to serve as input to the model. With 400 million images to ingest, image decoding represented a key computational bottleneck in the preprocessing stage and was parallelized across available CPU cores. Decoded images were then passed directly to the YOLO inference pipeline without additional transformations, leaving the Ultralytics Python package to internally handle resizing and padding to the target resolution.

### 3.1.4 Output

Each detected visual element was recorded along with bounding box coordinates, detection confidence score, and the source scan filename for provenance tracking.

## 3.2 Results

The detection step took approximately 16 days across all batches for an average of 22 minutes and 47 seconds per batch of 1,000 volumes on the machine described in Appendix G, with an average of 401 pages per volume. The bottleneck lay mainly in CPU- and I/O-bound preprocessing rather than GPU inference: 89.5% of time per scan was spent on image loading from disk, JPEG/TIFF decoding, resizing, and tensor normalization, while GPU inference with the YOLO model accounted for only 9.8%, with an additional 0.8% for GPU cache operations. Across the 394 million page scans processed, the CPUs had to sustain preprocessing for four concurrent GPU streams. Disk and network I/O for fetching cached image data also constrained the achievable feed rate. As a result, the four NVIDIA L40S GPUs we dedicated to this task were substantially underutilized: although each GPU could theoretically sustain ~152 scans/s, they were effectively driven at only ~71 scans/s, yielding an aggregate wall-clock throughput of 285 scans/s and only ~47% of the combined theoretical detection throughput, with performance dominated by data preparation rather than compute.

There were a total of 43,666,415 detections before post-hoc filtering (See Section 8.1). Upon reviewing samples from extractions across volumes, we set a confidence threshold cutoff (See Section 8.1). After applying this filter, 28,129,631 detections remained. After deduplication (See Section 5), 22,622,060 total detections remained across 983,004 volumes, of which 895,014 had at least one detection and 87,990 had none. The distribution of detections per volume was heavily right-skewed: although the average volume contained 25.3 detections, the median was only 4 detections. The most extreme case had 5,260 detections; manual inspection shows that this volume is a densely illustrated art reference work, containing thousands of images of Greek and Roman statuary (See Appendix B). This kind of richly illustrated catalogue explains the long tail at the high end of the distribution and indicates that a small number of highly visual works contribute disproportionately many detections, as seen in Figure 3. The average number of detections per volume is therefore inflated by these “Spiders Georg” volumes (“Spiders Georg,” 2013).

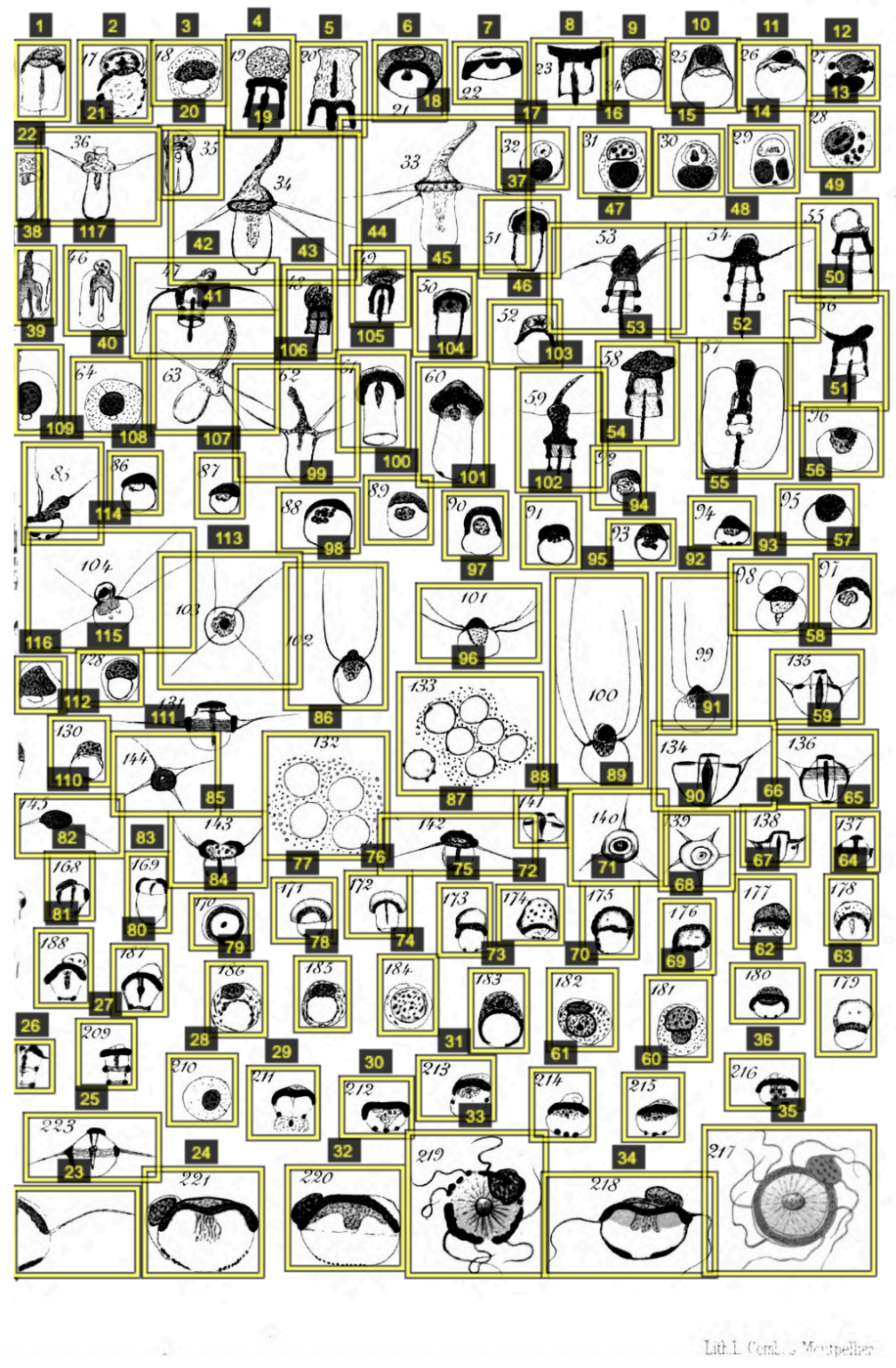

*Figure 3:* Screenshot.
Example of a volume from the collection with many detections in close proximity. Images such as these helped inform tuning the training thresholds, as seen from the VIA image annotator.

The resulting median bounding box size is 751×683 pixels, mean size of 904×843 pixels. At the extremes, crop sizes range from 37×21 pixels up to 10,107×8,471 pixels, with the largest regions typically capturing full-page or near-full-page illustrations found in art catalogues and similar works. Figure 4 depicts a scatter plot of the crop width vs crop height in pixels.

Future versions of the pipeline could explore using VLMs for detection or region proposal at scale. While large VLMs are currently too expensive for most libraries to run naively on hundreds of millions of pages, hybrid approaches such as using lightweight detectors to propose candidates and VLMs only for borderline cases, could yield better recall and more nuanced categorization without prohibitive cost.

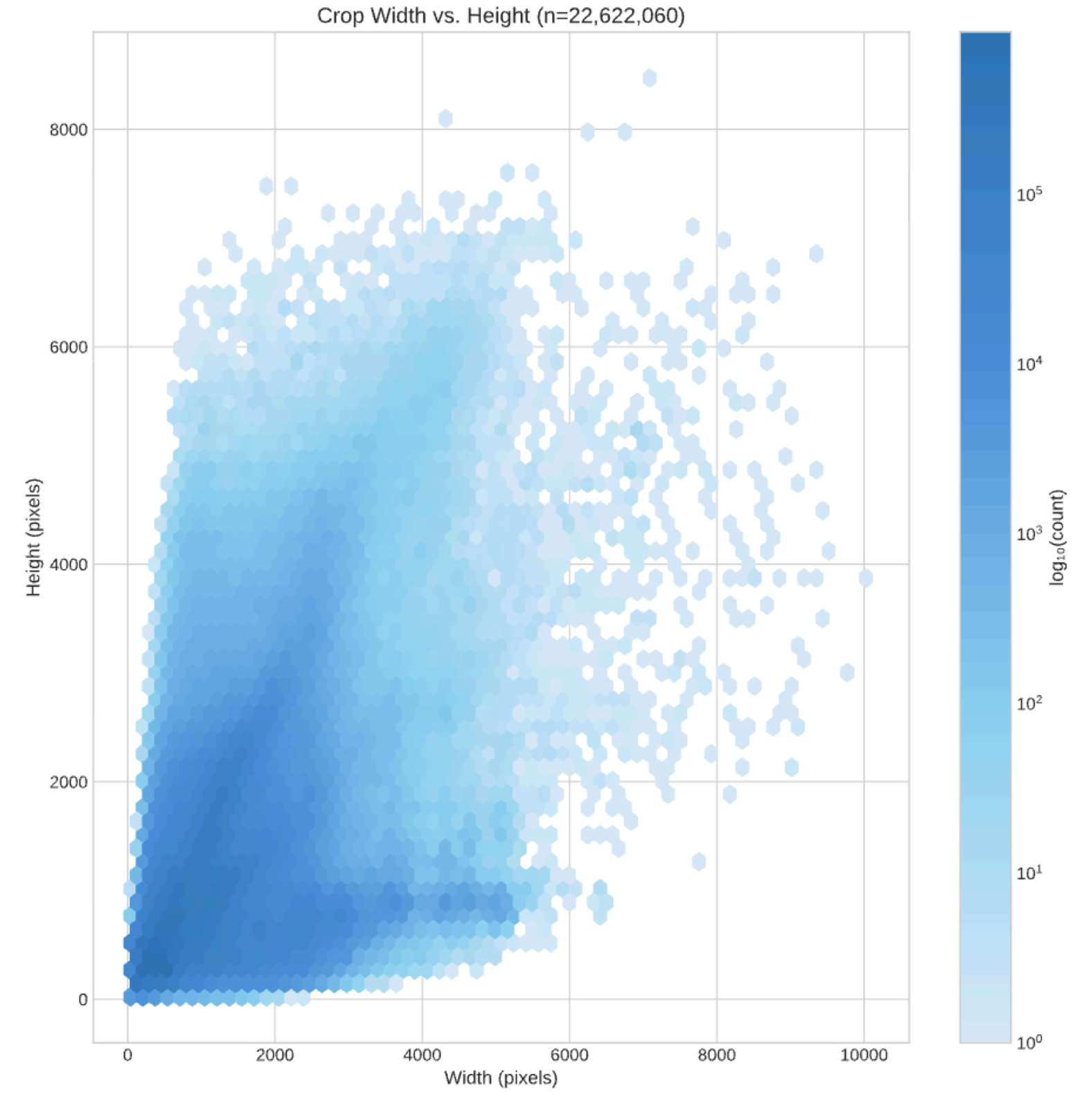

*Figure 4:* Scatter plot.
Detection crop width vs height in pixels.

# 4 Classification

## 4.1 Methodology

Following detection, the classification step categorizes each localized visual element into high-level, semantic classes. Classification operates exclusively on the cropped regions produced by the detection stage, therefore processing only volumes that contain at least one detected element. Classification employs a YOLO classification variant, YOLO26s-cls with 2.8 million parameters.

### 4.1.1 Training

The training set for this classification step consisted of 6,042 crops, 1,602 annotated manually, 4,440 pre-annotated using an intermediary model and manually edited. These crops come from the same source as the detection dataset, with each crop consisting of one specific element on a specific page, defined by bounding boxes from the detection training dataset. Training was done using the class set shown in Table 2. This set of classes was assembled through iterative testing and was designed to balance specificity, labeling burden, and model performance. For the purpose of training this model, classes were considered mutually exclusive.

The classification model was trained using an input resolution of 480×480 pixels, using a batch size of 32 crop images on 4 GPUs. The model trained for up to 275 epochs with a patience set to 100, with the best checkpoint obtained at epoch 140. Total training time was approximately 30 minutes. Mixed precision and deterministic training were both enabled.

*Table 2: Classification data splits by class*

| Class | Description | Train | Val | Test | Total | % of total samples |
|---|---|---|---|---|---|---|
| Image/Illustration | Photographs, drawings, engravings, and pictorial content | 2596 | 324 | 325 | 3245 | 53.70% |
| Music | Musical notation and scores | 1119 | 139 | 141 | 1399 | 23.20% |
| Artifact | Scanning errors, book clips, fingers, and other elements that are not part of the published work | 403 | 50 | 51 | 504 | 8.30% |
| Ex Libris/Decorative | Bookplates, ornamental borders, and decorative elements | 370 | 46 | 47 | 463 | 7.70% |
| Chart/Graph | Statistical visualizations, diagrams, and data representations | 344 | 43 | 44 | 431 | 7.10% |

Since the training set is limited in size, augmentations were used to improve generalization (Shorten and Khoshgoftaar, 2019). Color jittering in HSV space used a hue shift of 0.015, saturation of 0.7, and value of 0.4. Geometric augmentations included translation with a magnitude of 0.1 and scaling up to 0.5, along with horizontal flipping applied with a probability of 0.5. Mosaic augmentation was used at full strength (1.0) for training. RandAugment (Cubuk *et al.*, 2019) was enabled, and random erasing was applied with a probability of 0.4. The best model checkpoint yielded the per-class performance metrics shown in Table 3.

*Table 3: Classification model accuracy*

| Class | Precision | Recall | Support |
|---|---|---|---|
| Image/Illustration | 0.98 | 0.97 | 325 |
| Artifact | 0.98 | 0.99 | 51 |
| Ex Libris/Decorative | 0.96 | 0.96 | 47 |
| Music | 0.99 | 0.99 | 141 |
| Chart/Graph | 0.87 | 0.91 | 44 |

### 4.1.2 Inference

For inference, images were processed at an input resolution of 640 pixels with a batch size of 16 crops. As with detection, these values were chosen to increase throughput; no meaningful accuracy degradation was observed.

### 4.1.3 Preprocessing

For each volume, only the source scans containing detected elements were decoded from their compressed formats. This selective loading avoided unnecessary I/O for pages without visual elements. Each detection's bounding box coordinates were used to extract the corresponding region from the decoded scan. The crops were then converted to numpy arrays suitable for model input. Worker processes were distributed round-robin across available CUDA devices.

#### 4.1.4 Output

For each classified detection, we record the predicted class, the softmax probability for the predicted class, and the complete softmax vector across all classes. Storing the complete probability distribution enables post-hoc analysis of model uncertainty and supports alternative decision boundaries without re-running inference.

### 4.2 Results

Running the classification step on the machine described in Appendix G took a total of 3 days 17 hours across all batches, averaging 5 minutes 19 seconds per batch of 1,000 volumes to classify 43.7 million crops. For classification, 55.7% of per-crop time was spent in YOLO26s-cls inference on the GPUs. Because the inputs were small patches already derived from the detection output, the pipeline avoided the expensive JPEG/TIFF decoding and full-page resizing that dominated the detection step, leading to minimal disk I/O. As a result, the system sustained approximately 140 crops/s of aggregate throughput across all four GPUs, including pre-processing and GPU cache clearing. Each GPU averaged 75 crops/s during inference, making classification the only phase in which GPU compute was the primary limiting factor.

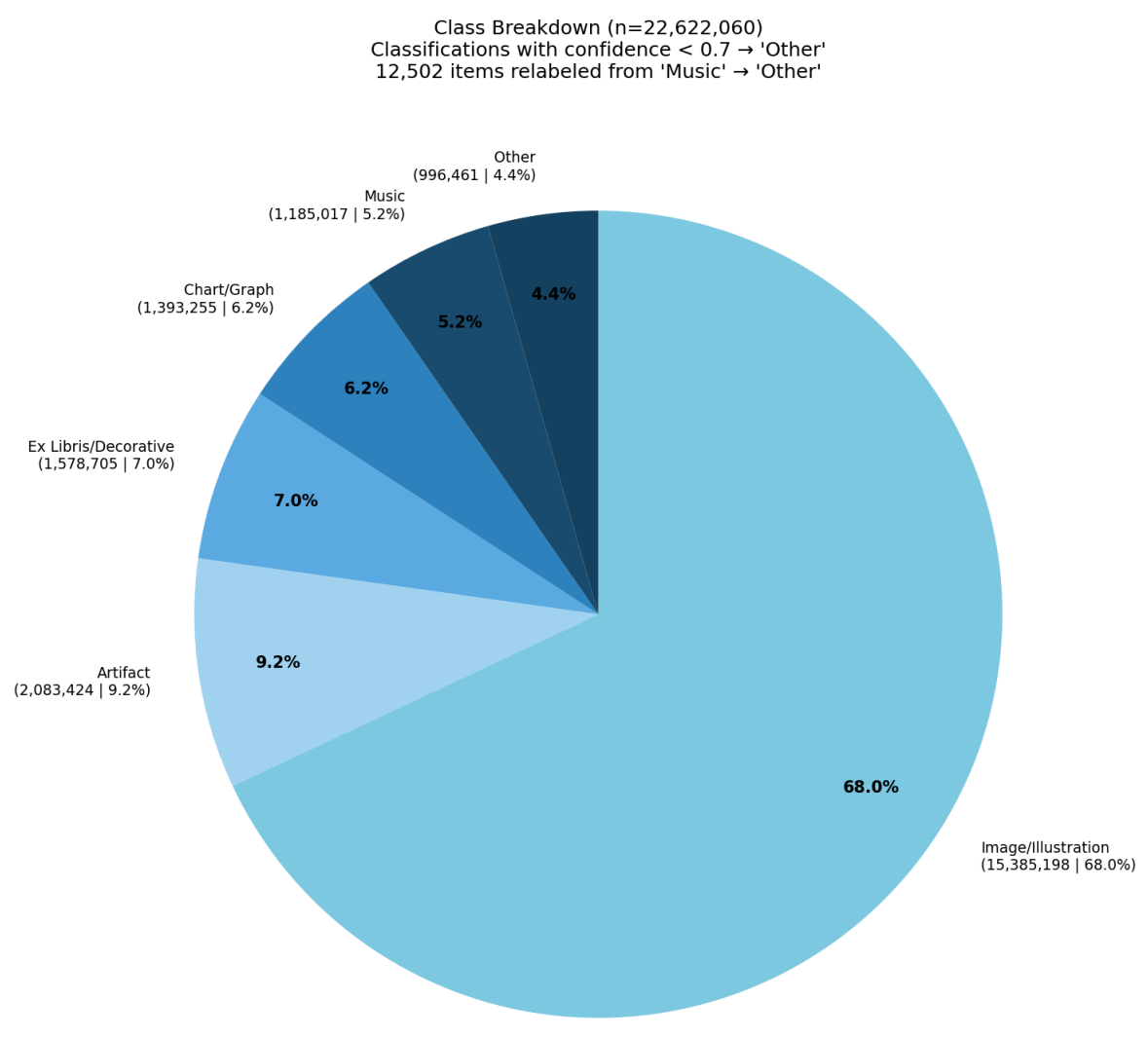

*Figure 5:* Pie chart.
Breakdown of detection crops after post-hoc filtering by classification category as labeled by our classification model.

After post-hoc filtering (see Section 8.1), 70.2% of regions were labeled as “Image/Illustration”, followed by 9.7% as “Artifact”, 8.0% as “Ex Libris/Decorative”, 6.8% as “Chart/Graph”, and 5.3% as “Music”. Relabeling predictions with low confidence as “Other” (See Section 8.1) shifted 983,959 crops, or 4.3% of the dataset, leaving 68.0% “Image/Illustration”, 9.2% “Artifact”, 7.0% “Ex Libris/Decorative”, 6.2% “Chart/Graph”, and 5.3% “Music”. The reclassification rates highlight which categories are most ambiguous to the model, with 12.9% of “Ex Libris/Decorative” and 9.1% of “Chart/Graph” being reclassified, suggesting that decorative bookplates, printer’s marks, and schematic figures are the hardest to separate cleanly.

Model confidence scores indicate that, for the vast majority of detections, the classifier was “extremely certain.” The average confidence was 0.966 with a median of 1.000, and 85.9% of all predictions lay in the 0.95–1.00 range, while only 0.3% fell below 0.50. Per-class statistics show the same pattern: “Music” and “Image/Illustration” have the highest average confidences with 0.994 and 0.974 average confidence scores respectively, whereas “Ex Libris/Decorative” is notably lower at 0.904. This may be due to various factors, such as low training data for these classes or weak inductive biases in decorative items. Figure 6 shows an example of crops that may introduce ambiguity.

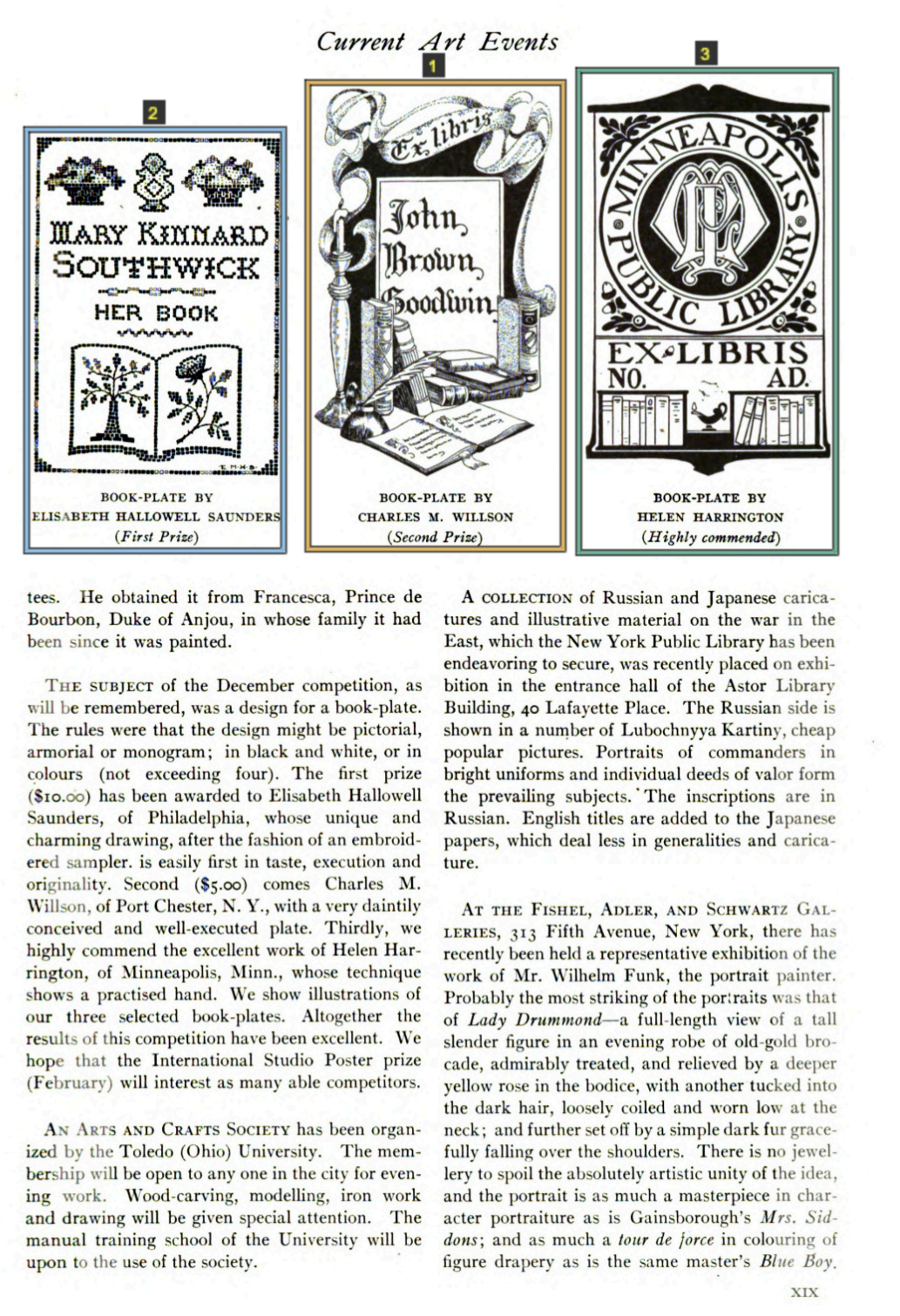
*Current Art Events*

BOOK-PLATE BY ELISABETH HALLOWELL SAUNDERS (*First Prize*)

BOOK-PLATE BY CHARLES M. WILLSON (*Second Prize*)

BOOK-PLATE BY HELEN HARRINGTON (*Highly commended*)

tees. He obtained it from Francesca, Prince de Bourbon, Duke of Anjou, in whose family it had been since it was painted.

THE SUBJECT of the December competition, as will be remembered, was a design for a book-plate. The rules were that the design might be pictorial, armorial or monogram; in black and white, or in colours (not exceeding four). The first prize ($10.00) has been awarded to Elisabeth Hallowell Saunders, of Philadelphia, whose unique and charming drawing, after the fashion of an embroidered sampler. is easily first in taste, execution and originality. Second ($5.00) comes Charles M. Willson, of Port Chester, N. Y., with a very daintily conceived and well-executed plate. Thirdly, we highly commend the excellent work of Helen Harrington, of Minneapolis, Minn., whose technique shows a practised hand. We show illustrations of our three selected book-plates. Altogether the results of this competition have been excellent. We hope that the International Studio Poster prize (February) will interest as many able competitors.

AN ARTS AND CRAFTS SOCIETY has been organized by the Toledo (Ohio) University. The membership will be open to any one in the city for evening work. Wood-carving, modelling, iron work and drawing will be given special attention. The manual training school of the University will be upon to the use of the society.

A COLLECTION of Russian and Japanese caricatures and illustrative material on the war in the East, which the New York Public Library has been endeavoring to secure, was recently placed on exhibition in the entrance hall of the Astor Library Building, 40 Lafayette Place. The Russian side is shown in a number of Lubochnyya Kartiny, cheap popular pictures. Portraits of commanders in bright uniforms and individual deeds of valor form the prevailing subjects. The inscriptions are in Russian. English titles are added to the Japanese papers, which deal less in generalities and caricature.

AT THE FISHEL, ADLER, AND SCHWARTZ GALLERIES, 313 Fifth Avenue, New York, there has recently been held a representative exhibition of the work of Mr. Wilhelm Funk, the portrait painter. Probably the most striking of the portraits was that of *Lady Drummond*—a full-length view of a tall slender figure in an evening robe of old-gold brocade, admirably treated, and relieved by a deeper yellow rose in the bodice, with another tucked into the dark hair, loosely coiled and worn low at the neck; and further set off by a simple dark fur gracefully falling over the shoulders. There is no jewellery to spoil the absolutely artistic unity of the idea, and the portrait is as much a masterpiece in character portraiture as is Gainsborough’s *Mrs. Siddons*; and as much a *tour de force* in colouring of figure drapery as is the same master’s *Blue Boy*.

XIX

*Figure 6:* Screenshot.
Images of Ex Libris designs in a book. As our system tries to classify both Ex-Libris and Image/Illustrations, this creates ambiguity for the model. Though the images are Ex Libris, we labeled them as Image/ Illustration based on context. From barcode 32044039107198.

The “Artifact” category represents a variety of different types of visual elements. This class includes, for example, hands captured in scans, metal clips that were used during the scanning process, barcodes, fractured pages, etc. Figure 7 shows some examples of the “Artifact” class in the dataset.

The classifier often misclassified crops that resembled sheet music as belonging to the “Music” class. To address this issue, we applied additional filtering. After reviewing a subset of the true and false positives, we observed that many items that were indeed music had very high classification confidence scores (≥ 0.99). There were exceptions, but the majority of correctly identified music samples had confidence scores close to 1. We also investigated using a simple keyword-based approach. Using the generated captions, we retained crops that contained at least one term from a pre-selected list of music-related keywords. The keywords were translated using Gemma 4 31B (Google DeepMind, 2026) into the top 20 volume languages for crops found in the “Music” class (See Appendix C). This approach had two main limitations: many captions were in different languages, requiring translation of the keyword list, and some music-related captions did not explicitly contain any music-related terms. To mitigate these issues, we adopted a combined strategy: we applied a blanket confidence threshold of 0.99, classifying all items with confidence above this threshold as music regardless of their captions. For items with confidence scores below the threshold, we retained only those whose captions contained at least one of the selected keywords. Out of the 1,197,519 crops that were originally labeled as “Music”, 12,502 items were relabeled to “Other”.

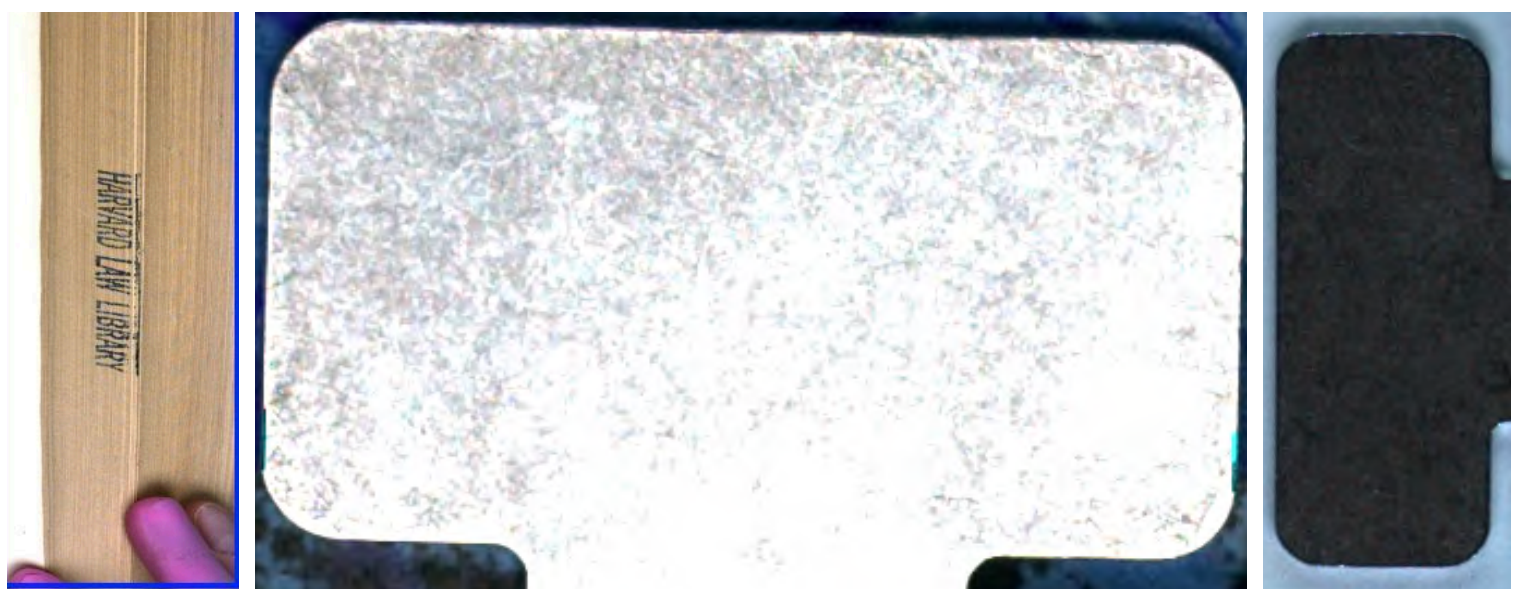

*Figure 7:* Screenshot.
Examples of the “Artifact” class in the dataset. Left: hand captured in page scan along with the page edges of the volume. Middle and right: clips used for holding pages down during the page scanning process.

# 5 Deduplication

## 5.1 Methodology

Large-scale digitization efforts inevitably produce duplicate or near-duplicative visual elements such as: repeated illustrations across volumes, decorative elements reused by publishers, identical pages appearing in multiple editions or human errors when scanning pages. To address these, our pipeline implements a two-stage deduplication strategy: image hashing for exact and near-exact matches, followed by semantic embedding similarity for visually-related but not identical content. Exact matches may be the result of scanning issues, such as pages scanned twice in a row and then included as-is in the digital copy of the volume. A more common scenario occurs when two images are highly similar, sometimes the same image but on different pages. In this case, due to slight visual differences that reduce the pixel level similarities between duplicates, image embeddings can be used to estimate similarity (Babenko *et al.*, 2014). More specifically, robustness for the perceptual hash algorithm may degrade under rotation beyond small angles, and under some geometric changes such as cropping and certain resizings (Monga and Evans, 2006; Zauner, 2010). Using both dense embeddings and sparse hashes enables a deduplication strategy where exact duplicates are identified cheaply via hash comparison, while content-level near-duplicates require the more expensive embedding similarity search.

### 5.1.1 Perceptual hashing

We computed a 144-bit perceptual hash (pHash) for each detection crop using the Python imagehash library (Buchner, 2026). We used a 12×12 DCT-based pHash, which produces a 144-bit binary code that captures the structural content of the image in a compact form suitable for efficient exact and near-duplicate detection.

### 5.1.2 Semantic embeddings

We generated dense vector representations using the Self-Supervised Copy Detection (SSCD) (Pizzi *et al.*, 2022) model. This 24.6 million parameter model uses a ResNet-50 backbone with GeM pooling and was specifically designed for copy detection tasks, producing embeddings that are robust to common image transformations. Specifically, we deployed the `sscd_disc_mixup` model, which was evaluated using the DISC2021 dataset (Douze *et al.*, 2022). Each detection crop was resized to 224×224 pixels and normalized using ImageNet statistics (Russakovsky *et al.*, 2015). The model operates on RGB inputs in CHW tensor format and outputs a 512-dimensional embedding for each crop. Embeddings are L2-normalized immediately after extraction, so that cosine similarity can be computed via inner products. Crops were processed in batches of 256 to maximize GPU utilization within memory constraints.

### 5.1.3 Model evaluation

To select a semantic embedding model, we constructed a small evaluation set of 376 images, converted to grayscale and organized into 70 duplicate groups, each containing between 2 and 13 near-duplicate items. Treating one representative from each group as unique yielded 232 unique images. We then compared five embedding models, four general-purpose multimodal encoders: CLIP (Radford *et al.*, 2021), OpenCLIP (Cherti *et al.*, 2023), MetaCLIP 2 (Chuang *et al.*, 2025), SigLIP 2 (Tschannen *et al.*, 2025) and one model specialized for copy detection: SSCD. For each model or hash configuration, we recorded precision, recall, F1, and embedding throughput on a single GPU, and selected the threshold that maximized F1 on this sample, as outlined in Table 4. For the hash-based methods the threshold was expressed as a maximum Hamming distance.

The specialized SSCD model provided the best balance of precision and recall among the learned embeddings. At its optimal threshold on this sample, SSCD achieved a precision of 0.93 and recall of 0.82, with throughput of roughly 123 images per second. General-purpose vision–language encoders performed noticeably worse on this near-duplicate detection task. OpenCLIP was the strongest of the four, with precision 0.72 and recall 0.76 at around 131 images per second, while CLIP and MetaCLIP 2 achieved lower F1 scores and SigLIP 2 performed worst overall. The pHash baseline was competitive in accuracy at higher hash sizes and substantially faster than any learned model, but its recall remained limited, particularly on noisy or heavily transformed scans, and it struggled with non-exact duplicates.

We experimented with PyTorch's DataLoader (Paszke *et al.*, 2019) to understand how different batching and shape-handling strategies affected both throughput and accuracy. Because DataLoader expects tensors of uniform size, we compared two approaches: resizing/cropping all images to a fixed resolution and padding images to a common size. Resizing all crops yielded an approximately 10× speedup, but reduced precision to 0.84 while leaving recall roughly unchanged at 0.82. In contrast, padding preserved more of the original content and increased precision to 0.92, at the cost of reduced recall (0.75) and slightly lower speed. Based on these results, our production pipeline used a simpler manual batching strategy: crops were resized to a fixed 224×224 resolution, stacked with `torch.stack`, and passed directly to the TorchScript model, preserving the speed gains of batched inference while keeping the implementation straightforward and leaving room for future tuning.

*Table 4: Deduplication model accuracy*

| Model | Precision | Recall | F1 | Threshold | Images/sec (1 GPU, batch=32) |
|---|---|---|---|---|---|
| SSCD | 0.93 | 0.82 | 0.87 | 0.83 | 123 |
| SigLIP 2 | 0.54 | 0.66 | 0.59 | 0.96 | 80 |
| OpenCLIP | 0.72 | 0.76 | 0.74 | 0.92 | 130 |
| MetaCLIP 2 | 0.52 | 0.72 | 0.61 | 0.96 | 31 |
| CLIP | 0.75 | 0.58 | 0.65 | 0.96 | 104 |
| *phash-size-8 | 0.77 | 0.44 | 0.56 | 14 | 289 |
| *phash-size-16 | 0.81 | 0.65 | 0.72 | 82 | 278 |
| *phash-size-24 | 0.80 | 0.64 | 0.71 | 222 | 268 |

*Here, we also include a perceptual hashing baseline (pHash) at three hash sizes 8, 16, and 24 where hash size n evaluates to an $n^2$-bit fingerprint. For the hash-based methods the threshold was expressed as a maximum Hamming distance.

### 5.1.4 Hash-based deduplication

The hash-based deduplication strategy we implemented in our pipeline operates directly on the 144-bit perceptual hashes. Given a user-specified Hamming threshold τ, two hashes are considered duplicates if their Hamming distance is at most τ (Manku, Jain and Sarma, 2007). We chose τ by optimizing for the best F1 score for our labeled duplication data.

Perceptual hashes were first exported from the database into a compact binary representation on disk, where each hash is packed into three 64-bit machine words. This representation enables word-level bitwise operations and efficient memory-mapped access for large collections.

### 5.1.5 Fuzzy matching via locality-sensitive hashing

To avoid the $O(n^2)$ cost of exhaustive pairwise comparison among $n$ perceptual hashes, we used a banding-style Locality-Sensitive Hashing (LSH) (Gionis, Indyk and Motwani, 1999; Leskovec, Rajaraman and Ullman, 2011) scheme tailored to fixed-length binary codes.

In this scheme, each 144-bit hash is partitioned into $B$ contiguous bands of $b$ bits each. For each hash h, we generate $B$ band keys of the form: $i \in \{0, \cdots, B - 1\}$ , $h \in \{0, 1\}^b$, where $h$ is the b-bit subvector corresponding to band $i$. For our setup, we chose $B = 6$ and $b = 24$.

We write out one record per hash-band pair and then externally sort this file by band index and band value using a disk-based sort. After sorting, any run of records sharing the same band index and band value corresponds to a bucket of candidate hashes that agree on that 24-bit band.

Buckets that are excessively large are skipped to maintain efficiency. For the remaining buckets, all contained hashes are treated as candidate near neighbors. For each bucket, all distinct pairs $(h_i, h_j)$ are enumerated and their exact Hamming distance computed via bitwise XOR and a popcount on the three 64-bit words:

$$d_H(h_i, h_j) = \text{popcount}(h_i \oplus h_j)$$

where $\oplus$ denotes bitwise XOR and popcount$(\cdot)$ counts the number of set bits. Pairs are retained only if:

$$d_H(h_i, h_j) \leq \tau$$

where $\tau = 16$ is the chosen Hamming threshold for approximate matches. These distance checks are parallelized across buckets using a process pool, with all workers sharing a memory-mapped view of the underlying hash array to avoid duplication of data in RAM.

In expectation, LSH reduces the average-case complexity from $O(n^2)$ pairwise comparisons to approximately $O(n)$, making large-scale fuzzy deduplication tractable. Specifically, on average, the size of each band-bucket is:

$$\frac{n}{2^b} = \frac{43.7M}{2^{24}} \approx 2.6$$

Thus instead of $n^2$ pairwise comparisons, we expect $2.6n$ many, apart from actual large classes of duplicates. The trade-off is controlled recall: some true matches may be missed if they never collide in any 24-bit band.

All verified similar pairs are then clustered into groups via the Union–Find (Cormen et al., 2009) (disjoint-set) data structure with path compression. Each connected component is assigned a canonical identifier, taken as the first in the group sorted by a unique identifier, ensuring deterministic and reproducible grouping.

### 5.1.6 Embedding-based deduplication

The second deduplication strategy of our pipeline operates on semantic embeddings to identify visually similar content that may differ enough at the pixel level that the pHash algorithm would produce hashes further in Hamming distance than our selected threshold. For example, two copies of the same illustration where one contains scanning-related variations in page curvature, differences in resolution, or different surrounding margins.

Our pipeline uses Facebook AI Similarity Search (FAISS) (Douze *et al.*, 2025) with the Hierarchical Navigable Small World (HNSW) algorithm for approximate nearest neighbor search, which provides sub-linear query time while maintaining high recall (Malkov and Yashunin, 2020). The HNSW index is configured with 12 connections per layer, a construction candidate list size of 100 when adding a node to the graph, and a search candidate list size of 30 for graph traversal during querying.

It employs `IndexHNSWFlat` with an inner product metric on L2-normalized embeddings. Since each embedding $x$ is normalized to $|x|_2 = 1$, the inner product

$$\langle x, y\rangle$$

is equivalent to cosine similarity, and the cosine distance is:

$$d_{cos}(x, y) = 1 - \langle x, y\rangle$$

The maximum cosine distance is set to a threshold of 0.14, which corresponds to a minimum cosine similarity of:

$$\langle x, y\rangle \geq 1 - 0.14 = 0.86$$

The index is built incrementally in batches of 100,000 embeddings to control memory usage during construction. Queries are executed in batches of 50,000 embeddings. For each embedding, up to $k = 30$ nearest neighbors are retrieved and then filtered to retain only those neighbors whose similarity satisfies $\langle x_i, x_j\rangle \geq 0.86$. To avoid redundant symmetric pairs, only pairs $(i, j)$ with $i < j$ are kept, yielding an edge list describing a graph of potential duplicates.

As in the hash-based stage, the Union–Find (disjoint-set) algorithm is applied to this graph to cluster embeddings into connected components.

### 5.1.7 Output

The results of both deduplication steps (hash-based and embedding-based deduplication) were written to dedicated tables, alongside an identifier for the deduplication group for downstream processing.

### 5.1.8 Design considerations

Both methods support HDF5-based caching (The HDF Group, 1997) of extracted hashes and embeddings. This enables rapid re-runs of collection-wide deduplication with different thresholds without re-extracting features from source images. Hash verification and embedding search are parallelized across available CPU cores. FAISS automatically utilizes OpenMP (Dagum and Menon, 1998) for HNSW operations. Embedding data is stored as a 2D dataset of shape (N, 512) in float32 format with gzip compression; HDF5 retains the array dimensions internally, so embeddings are indexed by row. Metadata for provenance tracking are stored as separate aligned 1D datasets in the same file. The HNSW index can optionally be persisted to disk for reuse across runs. In our run, the cosine distance threshold of 0.14 was empirically tuned to balance precision and recall. Hash-based deduplication with Hamming threshold 16 provides a more conservative strategy that catches only very close matches.

## 5.2 Results

The embedding and hashing step ran for a total of 1 day 23 hours across all batches on the machine described in Appendix G, averaging 2 minutes 48 seconds per batch of 1000 volumes to generate SSCD embeddings and pHashes for all crops. This stage achieved an effective throughput of 267 crops/s. This step reprocessed the same crops produced by classification, but with limited additional I/O since the patches were already resident. The 4 GPUs were evenly loaded and the step was GPU-bound, but efficient.

The deduplication stage started from 43.7 million candidate crops and used three complementary grouping strategies—hash-based, embedding-based, and their intersection—to identify and collapse near-duplicate content. Applied in isolation, perceptual hashing reduced the dataset by 22.2% to 33.96 million unique hash groups, while embedding-based clustering was more aggressive, reducing by 31.0% yielding 30.11 million groups, as expected since the embedding model can detect conceptually similar but not pixel-identical images.

For the final dataset, we chose a conservative strategy of deduplicating only on groups with agreement between both signals. A more efficient approach would have been to use the faster of the two deduplication methods, get a list of candidate duplicates, and then run the other method only on the duplicates. However, using the intersection was a downstream decision after the deduplication groups were created. Using this strategy, the hash/embedding intersection accounts for a 19.6% reduction to the final dataset. Group-size distributions show that most detections are unique regardless of the method: for the intersection, 94.4% of groups are singletons. At the same time, there are a handful of extremely large groups. Figure 8 shows examples from the two largest embedding deduplication groups before detection filtering (See Section 8.1): similar decorative items with 824,599 items, and barcodes with 418,436 items.

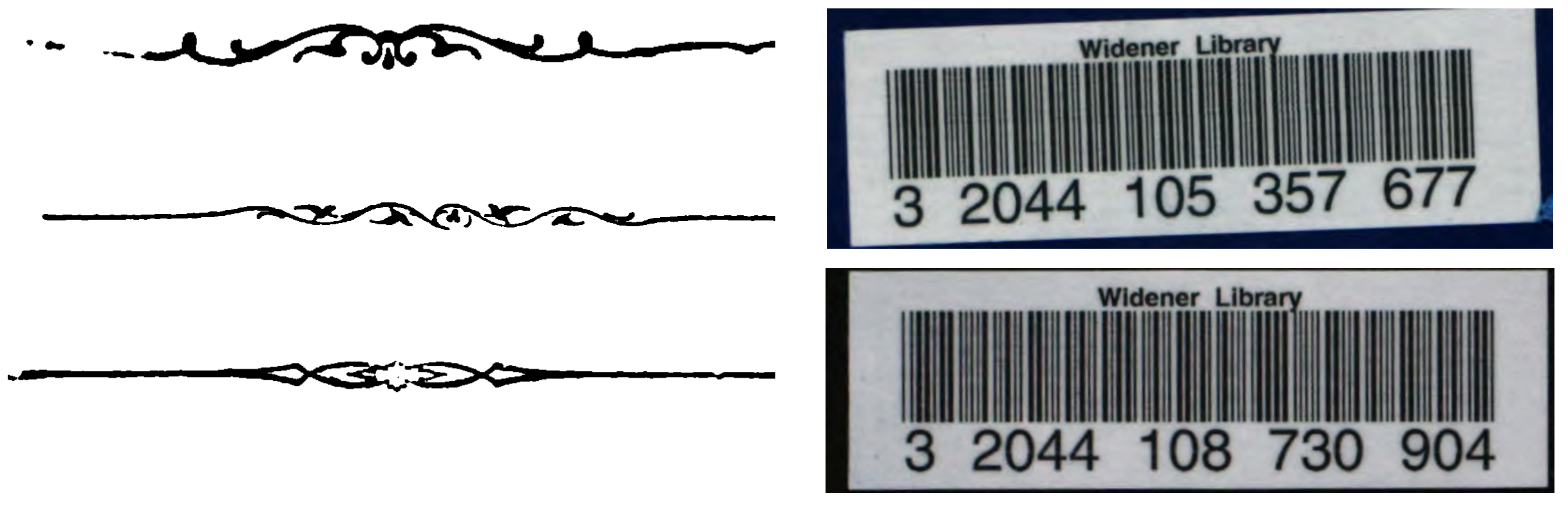

*Figure 8:* Screenshots.
The two largest embedding deduplication groups were decorative items and barcodes.

A direction for improvement to this step is the replacement or augmentation of the visual encoder with VLMs. Here, we used a visual encoder for embedding-based deduplication and a separate multimodal model for captioning. A natural extension would be to compute embeddings directly from a VLM trained or adapted for historical imagery, then reuse those embeddings both for semantic similarity search and for downstream tasks such as clustering, retrieval, or joint text-image analysis.

# 6 Experimental LLM-assisted captioning

## 6.1 Methodology

To create captions for eligible crops at scale, we used a multimodal large language model, as captions require both visual understanding capabilities as well as the ability to create accurate textual outputs (Lee *et al.*, 2025). Captioning using LLMs has been studied in the context of accessibility, search, and related vision-language tasks that jointly model visual inputs and natural language descriptions (Gubbi Mohanbabu and Pavel, 2024).

### 6.1.1 Experimental context

The captioning component represents an exploratory addition to the pipeline, designed to investigate the feasibility of generating descriptive metadata for visual elements at scale using large language models. Unlike the detection, classification, and deduplication stages, which employ well-established computer vision techniques that can be easily measured, automated captioning of historical visual content remains an open research problem with significant challenges around accuracy, hallucination, and domain specificity. We included this step to gather empirical data on LLM/VLM captioning behavior, and to explore the potential for LLM-assisted metadata enrichment in library contexts.

### 6.1.2 Approach

We employed OpenAI's GPT-4.1-nano (OpenAI *et al.*, 2024) to generate natural language descriptions for each detected visual element. The model received both the cropped image and surrounding page text as context, the goal of which was to help ground its descriptions in the document's content, as needed.

We identified early-on that not all visual element categories would benefit equally from textual captioning. Decorative elements, bookplates, and physical artifacts (stamps, seals) are largely self-describing or lack semantic content that would benefit from natural language description. We therefore excluded elements classified as "Ex Libris/Decorative" or "Artifact" from the captioning step, focusing computational resources on content-bearing imagery.

Each captioning request combined four information sources:

1. System Instruction: A prompt establishing the captioning task, constraints, and output format.
2. Page Text Context: OCR-derived text from the same page as the visual element, providing semantic grounding.
3. Image Crop: The detected region, resized to a maximum dimension of 1248 pixels, lightly compressed as JPEG and sent base64-encoded.
4. Language: The language that the caption should be written in, extracted from the Institutional Books dataset (`language_src`).

Figure 9 shows the final prompt used for extracting captions. Captions were generated in the source language of the volume when determinable from bibliographic metadata extracted from Institutional Books. Language codes were resolved to human-readable names via ISO 639 standards. When language metadata was unavailable or invalid, the system defaulted to English.

### 6.1.3 Log probabilities as a soft signal on captioning confidence

While not a direct measure of accuracy or confidence, log probabilities have been shown to be a useful proxy for plausibility and uncertainty (Jiang *et al.*, 2021). In the context of this experiment, we chose to capture token-level log probabilities and to compute average linear logprobs for all captions as a soft, coarse signal on the above. We hypothesize that this datapoint may help support downstream research on captioning accuracy and, more broadly, serve as an additional filtering criterion for direct uses.

### 6.1.4 Parallel processing architecture

Captioning was I/O-bound rather than compute-bound, with the majority of time spent waiting for data to reach and come back from OpenAI's API. Our implementation reflects this constraint. Worker processes were distributed across available CPUs, each handling a subset of volumes. Within each worker, a thread pool issued concurrent API requests with up to 8 simultaneous requests per batch. A module-level OpenAI client instance was reused across requests within each worker process to minimize connection overhead. Timeouts and retry logic were implemented for processing robustness (See Appendix D).

```
You are a librarian who writes precise, concise captions
for image crops.
- Use the page text only as supporting context, and only
when it clearly relates to what is visible in the image
crop.
- Focus on describing what is visually present in the
crop. Do not summarize or restate the page text.
 - Do not add background information, interpretations, or
educated guesses that are not directly shown in the image
crop.
- If you are unsure about something, describe it
generically (for example, "a person," "a diagram," "a
building") rather than guessing.
- The caption must be 50 words or less.
- Reply only with the caption, in {language}, with no
additional commentary.
```

*Figure 9: Captioning prompt.*

This prompt design reflects lessons learned from initial experiments where the model exhibited tendencies toward hallucination and over-interpretation of historical imagery, as well as over reliance on the page text rather than the image itself. The page content and crop image are fed into the model as content blocks of types `input_text`, and `input_image`, respectively.

### 6.1.5 Output

For each caption, we recorded caption text, the language that was passed to the model prompt, and serialized token-level log probabilities.

In addition to the caption texts, we collected some additional data related to captioning:

- For each caption, along with the language that we passed to the model in the prompt, we also used the Lingua Python (Stahl, 2024) package to detect the language the caption was written in. This downstream step was to check the efficacy of the model to produce the caption in the desired language. However, this comes with limitations. Lingua supports fewer languages than were previously identified in the Institutional Books: Harvard Library dataset. Additionally, it has been shown that language-identification accuracy decreases as text length decreases (Goldszmidt, Najork and Paparizos, 2013).
- A set of keyword matches from the Chronicling America Thesauri (Gilmore, 2022) on race, ethnicity, citizenship and immigration. These thesauri were preserved by Harvard Law School Library's Public Data Project (Harvard Library Innovation Lab, 2026) . For each caption, we included the count of how many exact matches were found for each keyword. The goal of this additional export was to help support future research, for example on how LLMs process historical materials. *These keyword matches are "naive" and should not be used directly to detect specific types of contents or infer text characteristics.*

### 6.1.6 Limitations and future directions

The experimental captioning component has several known limitations that inform its positioning as exploratory rather than ready-to-use data. Vision-language models may generate plausible-sounding but factually incorrect descriptions (Rohrbach *et al.*, 2019). We hypothesize that this is particularly true for ambiguous historical imagery. The logprobs data may provide a foundation for hallucination detection and adherence to task, but does not eliminate the underlying risk (Jiang *et al.*, 2021). We hypothesize that general-purpose vision-language models lack a sufficient amount of specialized knowledge of historical visual conventions, printing techniques, and period-specific iconography for this task. Furthermore, multilingual outputs introduce the potential for model error, where the model may produce outputs in the wrong target language or in a contextually inappropriate language (Chen *et al.*, 2025). Fine-tuning on domain-specific data may improve caption relevance (Zhou *et al.*, 2019). Assessing caption quality for historical imagery is inherently subjective and labor-intensive and we are exploring automated evaluation metrics and sampling-based human evaluation protocols.

In addition to accuracy, appropriateness is a concern. As part of this project, we tested pre-trained content safety models on these generated captions in order to detect potentially harmful patterns in the captioning process. From our initial results, we quickly identified that this endeavor needed to be a standalone research project. We hypothesize that the notions of content appropriateness and safety are deeply contextual, and that tooling needs to be designed specifically for historical collections. As such, the captions generated as part of this project are provided "as is".

The captioning step of our resulting pipeline is designed to be an optional and independently re-runnable processing step, allowing the pipeline to operate without captions when cost or quality concerns predominate. While we used OpenAI's API as part of this experiment, we designed this to be interoperable enough to generate those captions with other inference platforms and models.

## 6.2 Results

The captioning step ran for a total of 6 days 22 hours across all batches, averaging 9 minutes 56 seconds per batch on the machine described in Appendix G. It generated 28.6 million captions for 750,035 volumes, with a failure rate of 0.07%. After post-hoc filtering (See Section 8.1), the total caption count was 18,499,081 for a total of 766,992,447 `o200k_base` tokens (OpenAI, 2023). Because we made use of an external inference API, this step ran entirely without local GPUs. 128 CPU worker processes each maintained an OpenAI API client and issued requests in parallel, keeping the CPUs fully saturated while also maximizing concurrency against the OpenAI captioning endpoint. Effective throughput was 47.8 captions/s wall-clock, corresponding to about 2.68 CPU-seconds per caption and a total of ~21,261 CPU-hours consumed. Text token generation returned by the API accounted for only 2–3% of the end-to-end time, with the dominant cost instead arising from the remote vision encoder and model forward pass on the OpenAI side, plus local overhead for request construction, serialization of image crops, and HTTP I/O.

For the most part, the language of the returned caption reflected the language that was fed into the prompt. Using Python's Lingua package, we analyzed the match rate based on the language detected. Out of the 18.5 million captions, 1.5% of them did not match the language that was fed into the model (See Appendix E).

For items with captions in the resulting dataset, the number of tokens per caption has a mean of 41.5 and a median of 41.0, with a standard deviation of 14.0. The shortest caption contains 1 token, while the longest contains 103 tokens.

Using the per-token log-probabilities, at each position in the caption we took the log-probability that the model assigned to the token that was actually generated at that position. For a caption with $T$ tokens, we first computed the average log-probability per token, and then exponentiated this value to obtain the geometric mean per-token probability:

$$\exp\left(\frac{1}{T}\sum_{t=1}^{T}\log p_{\theta}(x_t)\right)$$

This quantity is equivalent to the inverse perplexity for that caption and summarizes how likely the model judged the entire token sequence to be, normalized per token (Hills and Anadkat, 2023). Figure 10 (bottom) reports this geometric mean per-token probability for the captions produced by the model.

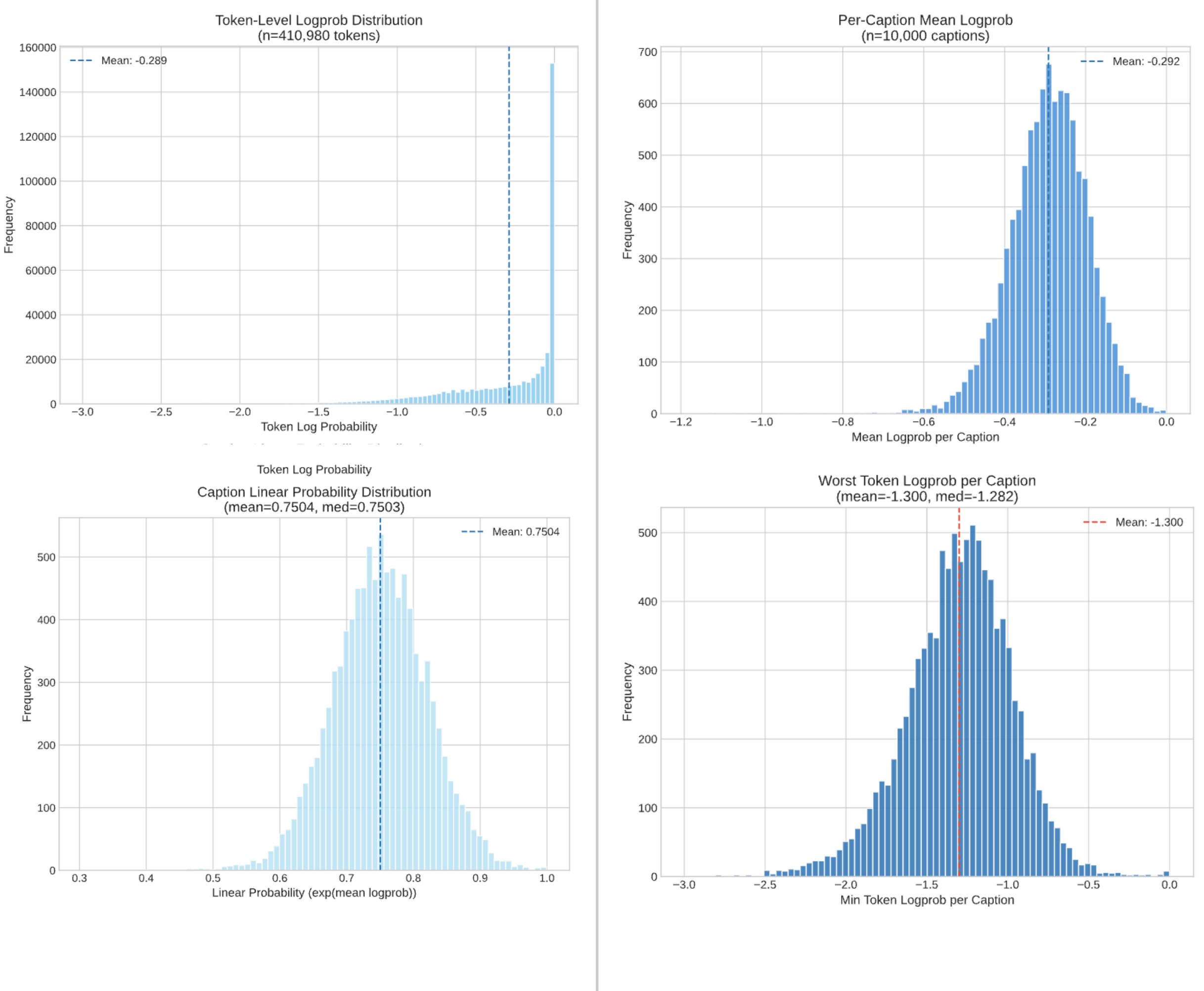

*Figure 10:* Histograms.
Top: Distribution of log-probabilities over all individual tokens for a sample of 10,000 captions.
Bottom: Distribution of average per-token probabilities for a sample of 10,000 captions.

*Figure 11:* Histograms.
Top: Distribution of the average log-probability of each caption's tokens for a sample of 10,000 captions.
Bottom: Distribution of the worst (most negative) token log-probability for a sample of 10,000 captions.

Although the prompt was constructed in a way to mitigate the model's reliance on the provided page text when the visual element was ambiguous, there were instances of the page text being summarized by the model instead of describing the visual element as seen in Figure 12. These cases, for the most part, were filtered out using the detection confidence threshold (See Section 8.1).

**Caption:** A person demonstrates the use of a soroban, a traditional Japanese abacus, with fingers on the device. The soroban features beads above and below a central beam, used for performing arithmetic calculations. The image emphasizes the hand positioning and the structure of the soroban.

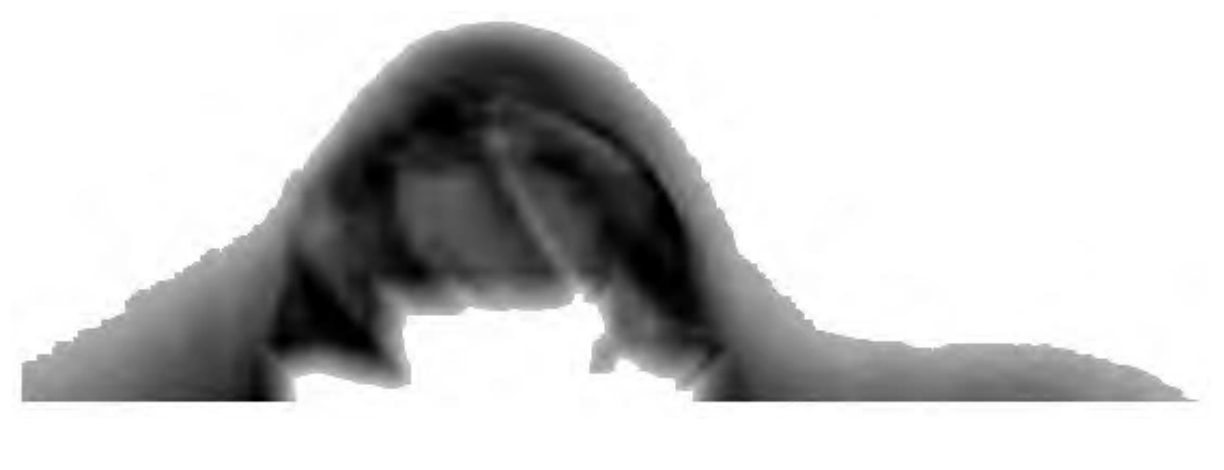

*Figure 12:* Screenshot.
When the model was unsure of how to describe an element due to a detection/classification error or a vague image, it relied heavily on the page context as seen here, where the caption describes the contents of the text rather than the image itself. In this case, the image was a classification error, where this should be in the "Artifact" class, as this is a finger that was captured during the scanning process. From barcode 32044000028530.

# 7 Orientation correction

## 7.1 Methodology

Correcting image orientation was an ad-hoc post-processing step that was not part of the original pipeline. We included this step to further improve the usability of the data. Automatic orientation correction is a well-studied problem in image processing, and a number of dedicated models and systems exist (O'Gorman, 1993; Lu, Wang and Tan, 2007). However, initial experiments with several off-the-shelf pretrained orientation and document-analysis models showed that their accuracy on scanned historical page imagery was insufficient for our purposes. We hypothesize that this was largely due to a domain gap, as historical document image datasets remain comparatively few, small, and inconsistent in format and annotation (Nikolaidou et al., 2022). Previous work on historical layout analysis has emphasized both the breadth of challenges posed by such material and the limited robustness of existing systems when moved beyond their target domains (Antonacopoulos et al., 2011). This limits the ability of generic orientation models to generalize to our setting. Furthermore, tight crops are less likely to contain visual anchors that may help established techniques detect the correct orientation. We see both our orientation model and the accompanying dataset as an opportunity to improve the usability of automatic orientation tools for historical collections.

We initially attempted to assemble a training dataset using a vision–language model (VLM) to infer page orientation from image–caption pairs. In practice, this approach proved unreliable: most VLMs we tested were not designed to infer fine-grained geometric properties such as rotation angle, and often produced internally inconsistent or ambiguous orientation cues. Consequently, we moved to a fully supervised formulation and trained a dedicated orientation classifier.

We framed orientation correction as a four-way classification problem, where the model predicted the rotation needed to restore an image to upright orientation. Images were first manually corrected to upright orientation using a small set of human-provided labels. Each upright image was then synthetically rotated by 0°, 90°, 180°, and 270° to yield four training samples per original image. The target label for each sample represented the correction (inverse rotation) required to bring the image back to the upright pose. This synthetic expansion produced a total of 31,616 samples from 7,904 unique crops, with the following split: 25,292 train samples, 3,160 validation samples, and 3,164 test samples. For each image, the synthetically rotated images are kept in the same split as the original to prevent contamination. The images for this training set were sampled from the deduplicated, filtered dataset (See Section 8.1).

The underlying label distribution in the original data was highly imbalanced: 93.0% upright, 5.6% rotated 90° clockwise, 1.3% rotated 90° counterclockwise, and 0.03% rotated 180°. This reflects the fact that most scanned pages were already upright. Synthetic rotations balanced the effective supervision signal across the four classes.

We used an EfficientNet-V2-M backbone (Tan and Le, 2021) with an input image size of 480px and a lightweight classification head consisting of a dropout layer followed by a linear layer mapping the 1,280-dimensional feature vector to four output logits. In total, the model comprises 52.8M parameters. During training, images were first resized to 512 × 512 px and then randomly cropped to 480 × 480 px. To improve robustness to contrast variations, partial occlusions, and scanner artifacts, we applied a set of moderate color and geometric augmentations. Specifically, random auto-contrast was applied with probability 0.3, random inversion with probability 0.15, random conversion to grayscale with probability 0.2, and Gaussian blur was applied to every training image. In addition, random erasing was used with probability 0.2.

For validation and test, we used a deterministic preprocessing pipeline: resize to 512 × 512 px, center crop to 480 × 480 px, and normalization with ImageNet statistics. We trained the model for up to 20 epochs with a batch size of 32, using a single NVIDIA GH200 GPU. Training converged rapidly, with both training and validation accuracies improving steadily over the first 10–12 epochs. See Appendix F for training and validation accuracy and loss across epochs.

The model achieved high training accuracy, while validation accuracy plateaued. The steady rise in validation accuracy accompanied by a slowly increasing validation loss after epoch 10 suggested moderate overfitting; however, the heavy augmentations appeared to regularize the model sufficiently that generalization to the held-out test set remained strong. On a single GH200 GPU, the full 20-epoch run took 58 minutes and 13 seconds. With 25,292 training samples per epoch and a batch size of 32, this corresponded to a throughput of approximately 145 images per second.

On the held-out test set of 3,164 samples, the model achieved an overall accuracy of 91.34%. Per-class accuracies are shown in Table 5.

*Table 5: Accuracy of orientation correction model per class on test set*

| Class | Accuracy % | Correct/Total |
|---|---|---|
| upright | 91.78% | 726 / 791 |
| rotate_90_clockwise | 89.76% | 710 / 791 |
| rotate_180 | 91.91% | 727 / 791 |
| rotate_90_counterclockwise | 91.91% | 727 / 791 |
| **Overall** | **91.34%** | **2890 / 3164** |

The model performed consistently well across all four classes, with slightly lower accuracy on the 90° clockwise class but no severe degradation on any orientation. Among the 791 test images evaluated for detailed error analysis, 65 predictions corresponded to incorrect corrections. Qualitatively, these errors were often associated with challenging or ambiguous cases, such as pages with predominantly graphical content, heavy degradation, or near-symmetric layouts.

### 7.2 Results

For inference, the orientation correction was only applied to the "Image/Illustration" and "Chart/Graph" classes. Inference time for running this model on the 16.7 million eligible images took a total of 2 days 23 hours on a GH200 GPU for an average of 65 images/second. Images were passed to the GPU in batches of 64 images, resized to 512 × 512 px and center cropped to 480 × 480 px.

Because the vast majority of pages in the full corpus were already upright, false positives were more costly in practice than false negatives. To reflect this asymmetry, we applied an additional confidence-based filter at deployment time. Based on manual review of out-of-sample inference results, we set a confidence threshold of 0.99 on the predicted class probability. Images with a maximum class probability below this threshold were left unchanged, whereas images with a maximum probability ≥ 0.99 were rotated according to the model's prediction. This conservative policy reduced the number of spurious corrections on already-upright pages while still automatically fixing the majority of genuinely mis-rotated images. Using this cutoff, 1,612,624 were rotated out of 16.7 million eligible images.

## 8 Dataset preparation

### 8.1 Methodology

At every step of the pipeline, samples were manually reviewed and checked for accuracy, robustness, and edge cases. This manual validation helped stress test weaknesses in the models and pipeline steps.

The final dataset preparation turned all intermediate pipeline outputs into a single, deduplicated dataset of visual elements. We started from all detections produced by the detection model and retained only those with confidence ≥ 0.75. This stricter downstream cutoff was chosen after visual inspection of borderline crops to remove obvious false positives while preserving relevant visual content. For each remaining detection, we applied a second confidence threshold at the classification stage. If the classifier's top-1 confidence was below 0.70, we relabeled the element to an "Other" catch-all class rather than forcing assignment to one of the five semantic categories.

Deduplication was applied conservatively by intersecting both the embedding and hash groupings (See Section 5). We chose to only collapse detections that shared both a hash group and an embedding group, with all other detections being treated as unique. Within each intersected group, we selected a single representative: the detection with the highest detection confidence, with ties broken deterministically. This yielded one canonical crop per strong duplicate cluster while preserving isolated or ambiguous cases.

For each remaining detection, we computed simple geometric features such as crop width, height, and pixel area. Each final record combines information from all of the pipeline steps, as well as the backfilled data from the captioning process (See Section 6.1.5). The final dataset structure can be seen in Appendix A. For each column, a suffix of either `_src`, `_gen`, and `_exp` is applied (See Appendix A). This is meant to delineate source, generated, and experimental variables. Source variables, labeled `_src`, include any data from the source dataset and metadata extracted from the original Institutional Books: Harvard Library dataset. Generated variables, labeled `_gen`, describe variables that were computationally derived from the source material, whether through AI models or other computational methods. Finally, the `_exp` suffix describes variables that are part of an experimental portion of the pipeline. In this case, the LLM captioning represents an experimental process, and is labeled as such.

## 8.2 Results

The final Institutional Books Visual Elements dataset contains 22,622,060 deduplicated visual elements drawn from the 983,004 volumes that comprise the Institutional Books: Harvard Library dataset. To understand how this visual subset relates to the broader text-only corpus analyzed in prior work, we compared their temporal, linguistic, and topical distributions. These comparative analyses were used to evaluate and refine the assumptions that guided the initial design of the project. While several patterns aligned with our prior expectations, it was essential to empirically verify them through systematic data analysis.

Overall, the visual dataset preserves the broad shape of the underlying collection but is measurably skewed toward more recent, English-language, and visually intensive subject areas. The dataset is therefore strongly concentrated in the nineteenth and twentieth centuries. While around two-thirds of the volumes in the text-only dataset were published between 1800 and 1999, the visual elements dataset is even more focused on this period (See Appendix H). Approximately 64% of volumes with at least 10 detections come from the 1800s and 33% from the 1900s, compared to 44% and 23% of volumes in the text-only corpus, respectively. Earlier centuries are sparsely represented in both datasets, and are effectively negligible in the visual subset. This temporal skew aligns with the history of book production: the nineteenth century marks a substantial expansion in the use of wood engraving, lithography, and later photomechanical processes, all of which increased the prevalence of illustrations in printed works (Gaskell, 1972).

Linguistically, English, German, and French remain the most represented languages in both datasets, but the visual collection is more heavily concentrated around these languages. Measured at the volume level, English accounts for roughly 47% of the text-only corpus, whereas English-language material represents about 60% of visual elements among volumes with at least ten detections. German and French are represented at similar rates in both datasets, while Italian, Spanish, Latin, and Russian appear proportionally less often among captioned elements than among volumes (See Appendix I). This pattern suggests that English-language books in the collection tend to be more image-dense than non-English counterparts, and that older, non-English works often predate the widespread adoption of illustrative printing and contribute fewer visual elements, even when present in the text corpus. Some of this imbalance may also reflect the collecting practices of the source institution—Harvard Library—while others may be due to the pipeline itself. Failures in detection may disproportionately reduce representation for works in languages whose pages may have different layouts that are harder for the detection model.

Topically, the visual dataset largely mirrors the high-level structure of the text-only corpus but, expectedly, exhibits a marked shift toward disciplines that traditionally rely on images. In volumes with at least 10 detections, “Science” and “Language & Literature” remain the top two categories in both datasets. While “Language & Literature” decreases its share compared to the textual dataset, “Science” more than doubles its textual counterpart. By contrast, “Law” is significantly underrepresented: it accounts for roughly 13% of the text-only collection but only 4.2% of volumes contributing visual elements. Several visually intensive disciplines are, by nature, substantially over-represented. “Fine Arts” volumes contribute over double their textual share, with the collection containing art historical monographs, catalogues raisonnés, and similar image-rich works. “Technology”, “Military Science”, “Naval Science”, and the “History of the Americas” all show increases in relative share, driven by technical diagrams, engineering schematics, maps, battle plans, and other diagrammatic content (See Appendix J).

The interaction between language and topic composition may partly explain observed gaps in language representation. For example, German-language volumes, which are disproportionately concentrated in “Philosophy, Psychology & Religion” (16.4% of German texts vs. 12.1% of English), reach the 10-detection threshold at only 49.3% compared to 53.9% for English. The effect is most pronounced for Latin, where 72.4% of volumes fall into “Language & Literature” or “Philosophy, Psychology & Religion” and only 36.8% reach the threshold.

These comparisons highlight that the resulting dataset is a systematically biased subset shaped by historical printing practices and disciplinary conventions. It amplifies the parts of the collection where images are structurally important such as art, technology, historical and military documentation, while filtering out domains such as law and some older humanistic texts that rely almost entirely on prose. For downstream users, this means that analyses or models trained on the visual dataset will over-reflect the visual cultures of the nineteenth and early twentieth centuries, English-language publishing, and image-heavy disciplines, and should be interpreted with this context in mind.

# 9 Pipeline footprint and optimization

The resulting pipeline was designed with two complementary goals: (1) to be runnable end-to-end on relatively modest hardware, within reach for most knowledge institutions, and (2) to scale horizontally as more compute becomes available. The footprint we report therefore covers both resource usage and operational characteristics across a full run of this pipeline against the source collection. For reference, we also included a smaller comparative run on a different, more constrained hardware environment.

We report here the aggregate computational footprint of running the full visual elements pipeline over the source collection on a multi-GPU server, as well as a smaller comparative run on an NVIDIA DGX Spark—a consumer-grade AI development system-on-chip. These measurements are intended both to document the practical cost of reproducing our results and to inform future optimizations or adaptations for different hardware environments. Note that orientation correction is not included in this analysis as it was done after the pipeline run and on a different machine.

The Institutional Books: Harvard Library collection comprises 983,004 volumes and 394,338,216 scanned pages, totaling 60 TB of data. These scans were processed locally on a GPU node, described in Appendix G. For GPU acceleration, we used four of the eight NVIDIA L40S GPUs available on this server node, each equipped with 46 GiB of VRAM. To avoid out-of-memory (OOM) conditions, we limited concurrent workers and tuned batch sizes so that per-GPU memory stayed within capacity. Due to RAM constraints and total collection size, we worked in batches of up to 1,000 volumes, streaming data through every processing step rather than loading larger portions of the entire collection at once. When data was sent and fetched over the network in later steps of the pipeline, effective link bandwidth became the bottleneck.

On the primary multi-GPU system, the complete pipeline processed 984 batches of up to 1,000 volumes over a span of 52 days. Of this, 41 days 17 hours were spent with the pipeline actively running, and 10 days 2 hours constituted downtime due to maintenance, reconfiguration, and restarts.

Within the active execution time, detection dominated the compute budget. The YOLO-based detection stage accounted for roughly 38% of active time (16 days). The next largest component was a dedicated caching step, which transformed and staged intermediate artifacts for later stages; this consumed approximately 20% of active time. Captioning, which relied on remote multimodal inference via API calls rather than local GPUs, represented 17% of active time. Classification consumed about 9% of active time. Storing final outputs accounted for roughly 10% with similar per-batch durations. Embedding and hashing together contributed 5% of active time, while the global deduplication pass—run once after all batches were processed—completed in roughly 7.5 hours, or less than 1% of the total active time. Across all steps, per-batch runtimes varied widely: batches with few detections completed in minutes, whereas those containing heavily illustrated volumes with thousands of large crops could drive per-step times into the one- to two-hour range. Table 6 shows the breakdown of the pipeline run by each step's duration.

Most interruptions were triggered by operational issues that were uncovered as the pipeline encountered corner cases, such as volumes with many unexpectedly large crops causing out-of-memory failures. In response, we iteratively refined resource limits and parallelism parameters, particularly the number of concurrent workers and per-GPU batch sizes in the most demanding steps.

To gauge how the pipeline might behave in a more distributed environment, we also executed a smaller run on an NVIDIA DGX Spark, processing nine batches end-to-end. This run completed in roughly 23 hours 8 minutes, with per-batch runtimes between 2 hours 16 minutes and 2 hours 44 minutes. The relative contributions of each step were broadly similar to the single-node case: detection again dominated at just over half of the total pipeline time (around 52%), captioning accounted for roughly 24%, and caching, classification, embedding/hashing, and storing each occupied between 4% and 9%. Detection remained the principal bottleneck, with average per-batch detection times around 1 hour 20 minutes, whereas other stages typically finished within 5–15 minutes per batch. Slight drift toward longer runtimes in later batches likely reflects differences in the underlying volumes rather than systematic slow-down, further emphasizing the impact of item-level variation on aggregate performance. Important to note here is that this run used a local Postgres database rather than a hosted one as in the full pipeline run, where a hosted database would have added some latency to the run.

*Table 6: Breakdown of step duration as part of the principal pipeline run, % of total span including downtime*

| Step | Hours | Percentage |
|---|---|---|
| Caching | 204 | 16.4% |
| Detection | 385 | 31.0% |
| Classification | 89 | 7.2% |
| Embeddings/Hashes | 47 | 3.8% |
| Captioning | 166 | 13.4% |
| Storing | 98 | 7.9% |
| Deduplication | 8 | 0.6% |
| Orchestration and Step-Transition Overhead | 3.5 | 0.3% |
| Downtime / Maintenance | 242 | 19.5% |

Overall, these measurements indicate that the full pipeline is tractable on an appropriately sized GPU server over a period of several weeks, and that it scales naturally to more parallel environments. They also highlight where future engineering effort is likely to be most impactful: reducing CPU-side preprocessing and I/O in the detection step, managing memory pressure for extreme volumes, and, where feasible, exploiting more aggressive batching and locality in the captioning and storage stages. See Appendix G for compute specs.

For captioning, the pipeline sent a total of 66.994 billion input tokens and 34.263 million requests to the OpenAI API. The total cost using the OpenAI API was $4,441.42. The cost was offset by prompt caching, since the structure of the prompt was the same across captions.

## 10 Rights determination

We respect the intellectual property rights of authors, publishers, and other rights holders. While we have taken deliberate steps to include only those volumes for which there is no known copyright restriction, specifically those identified by the HathiTrust Digital Library with a status of “public domain,” “public domain in the United States,” or “CC-Zero,” copyright determinations are complex and context-dependent, and occasionally subject to error.

While this is relatively low risk, some volumes in this dataset may be in the public domain in the United States but still subject to copyright or other rights protections in other jurisdictions. Additionally, the absence of an explicit copyright claim or rights status does not guarantee that a work is in the public domain, either in the U.S. or abroad. Information about the copyright status of individual volumes is provided on a good-faith basis and reflects available data at the time of determination, but we cannot guarantee its completeness or accuracy.

Users of this dataset will be solely responsible for making independent legal assessments about how and where they use the materials. Some uses of materials may also be restricted by trademark, privacy, publicity rights, or other such rights or restrictions. It is the user's sole responsibility to consider the possibility that such rights or restrictions may be involved and to secure any needed permissions. If any rights holder believes that a work included in this release is misidentified or improperly included, we welcome contact and will promptly review any concerns. Our goal is to provide broad public access while maintaining respect for intellectual property rights and ensuring responsible data stewardship.

## 11 Future directions

We view this release as a first iteration rather than a finished product. The pipeline has been engineered to be re-runnable in whole or in part: thresholds can be changed, models swapped out, and different types of visual elements can be extracted, all while reusing the existing pipeline structure. We hope that this flexibility will encourage others to experiment with alternative models and settings, contribute improved components, and build derived datasets and tools. This work is part of a broader effort to strengthen collaborations among collecting institutions, balancing the multimodal AI community's need for training data with institutional stewardship of digitized collections.

## Acknowledgements

This work was supported by unrestricted funding from Microsoft, OpenAI, Meta, and Jane Street as well as compute credits from OpenAI, which were used for image captioning.

AI tools were used to assist in the preparation of this technical report.

## Disclaimers

### Harmful Language and Content in this Dataset

This dataset is a collection of historical works that reflect the language, imagery, culture, and perspectives of their time. Users should be aware that some materials may contain language or portrayals that are outdated, offensive, or harmful today, such as racism, sexism, colonial attitudes, and other forms of discrimination. Some content may include inaccurate information, providing insight into historical contexts that existed at the time of writing. The materials are maintained in their original form to retain contextual understanding and facilitate research efforts, but we encourage critical awareness and cultural sensitivity for the creators and/or subjects of the collection. These materials are offered as part of a historical perspective, but should not be considered a stand-alone research collection constructed to give a balanced perspective on any topic.

### Harmful Language in Bibliographic Description

Metadata for this collection may contain language that is overtly or implicitly harmful, outdated, or biased, or may by omission fail to represent important perspectives. Metadata may contain language created decades ago. It is common practice within the field of library science to reuse descriptions provided from the creator of the materials. While in some instances this allows communities and individuals to represent their materials in their own words, unexamined use of this practice may mean that racist or other offensive terminologies appear in our description. We also use national standardized terms in our work that can be outdated and harmful. Note that terminology in historical materials and in library descriptions does not always match the language we currently understand to be preferred by members of the communities depicted.

Furthermore, we acknowledge that the act of collecting materials is not always neutral, and the work of describing and classifying library materials is influenced by inherent personal, institutional, and societal biases. Outdated or offensive terminologies may be present in metadata such as subject headings, and harmful language or bias may be introduced by catalogers supplying titles and descriptions. In other cases, books themselves present racist, offensive or otherwise harmful viewpoints in titles or descriptions that are routinely transcribed by catalogers.

**Note:** Some language in this statement was adopted from Harvard Library's statement on Harmful Language in Library collections[1].

## Generated and Experimental Content

This dataset contains generated and/or experimental content. While reasonable care was taken to ensure its quality, it is provided "as is," without warranties of any kind. It may contain errors or inaccuracies; users should verify the data independently and apply their own judgment.

[1] https://library.harvard.edu/harmful-language-library-collections

# Appendices

## Appendix A: Dataset fields

*Table App. A1: Field suffixes glossary*

| Suffix | Description |
|---|---|
| _src | *"From source"*.<br>This field's data comes from information we gathered from the collection itself. |
| _gen | *"Generated"*.<br>This field's data was generated as part of our analysis / post-processing. |
| _exp | *"Experimental"*.<br>This field's data was generated by downstream experimental models |

*Table App. A2: Row-level fields list*

| Field name | Type | Description |
|---|---|---|
| id | int | Unique integer identifier for the visual element. |
| crop_gen | bytes | WEBP export of the detected region. |
| barcode_src | string | Volume-level identifier or barcode for the source book. |
| page_filename_src | string | Filename of the source page scan within the volume. |
| bbox_xyxy_gen | list[float] | Bounding box of the detection in [x_min, y_min, x_max, y_max] pixel coordinates relative to the source page. |
| width_gen | int | Width of the crop in pixels. |
| height_gen | int | Height of the crop in pixels. |
| pixel_count_mpx_gen | float | Crop area in megapixels, (width_gen × height_gen) / 1,000,000. |
| detection_confidence_gen | float | Detection model confidence score for the element. |
| classification_gen | string | Final class label after post-processing. One of: Image/Illustration, Music, Artifact, Ex Libris/Decorative, Chart/Graph, or Other. |
| classification_confidence_gen | float | Confidence score of the predicted class before any relabeling to Other. |
| classification_probs_gen | list[object] | Full softmax distribution over classes, as a list of objects, sorted by prob descending. |
| phash_gen | string | 144-bit perceptual hash of the crop, serialized as a hexadecimal string. |
| embedding_gen | list[float] | 512-dimensional SSCD visual embedding (L2-normalized), stored as a list of floating-point values. |
| caption_exp | string | Caption text generated by a VLM (OpenAI's GPT-4.1-nano). Set to "CAPTION FAILED" when captioning did not succeed, or null for non-captioned elements. |
| caption_linear_prob_exp | float | Scalar confidence estimate derived from token-level log probabilities for the caption. |
| caption_lang_passed_exp | string | ISO 639-3 language code passed to the captioning model (intended caption language). |
| caption_lang_detected_exp | string | ISO 639-3 language code detected from the generated caption text. |
| caption_chronam_thesauri_matches_exp | object | Nested map of keyword matches from the Race, Ethnicity, Citizenship and Immigration Keyword Thesauri for Chronicling America; for each keyword, stores counts of matched surface forms. May be used for downstream research on VLM captioning of historical content (see Section 6). These keyword matches are "naive" and should not be used directly to detect specific types of contents or infer text characteristics. |
| orientation_correction_gen | string | Predicted orientation correction for the crop. Only populated for "Image/Illustration" and "Chart/Graph" classes. Predictions below the 0.99 confidence threshold default to "upright". |
| orientation_correction_confidence_gen | float | Max softmax probability from the orientation model. |
| orientation_correction_probs_gen | list[object] | Full 4-class orientation probability distribution, as a list of objects, sorted by prob descending. |

*Table App. A3: Fields nested under classification_probs_gen*

| Field name | Type | Description |
|---|---|---|
| label | string | Class label corresponding to a candidate semantic category. |
| prob | float | Softmax probability assigned to label. Entries are sorted in descending order of prob. |

*Table App. A4: Fields nested under caption_chronam_thesauri_matches_exp*

| Field name | Type | Description |
|---|---|---|
| top-level key (<keyword>) | string | Keyword from the thesauri |
| value | object | Map from surface form to count for this keyword. |
| nested key (<surface_form>) | string | Surface form of the keyword as observed in the caption text. |
| nested value (count) | int | Number of times this surface form was detected for the given keyword. |

*Table App. A5: Fields nested under orientation_correction_probs_gen*

| Field name | Type | Description |
|---|---|---|
| label | string | Orientation class label. One of: upright, rotate_90_clockwise, rotate_180, rotate_90_counterclockwise. |
| prob | float | Softmax probability assigned to label. Entries are sorted in descending order of prob. |

# Appendix B: Detection frequency

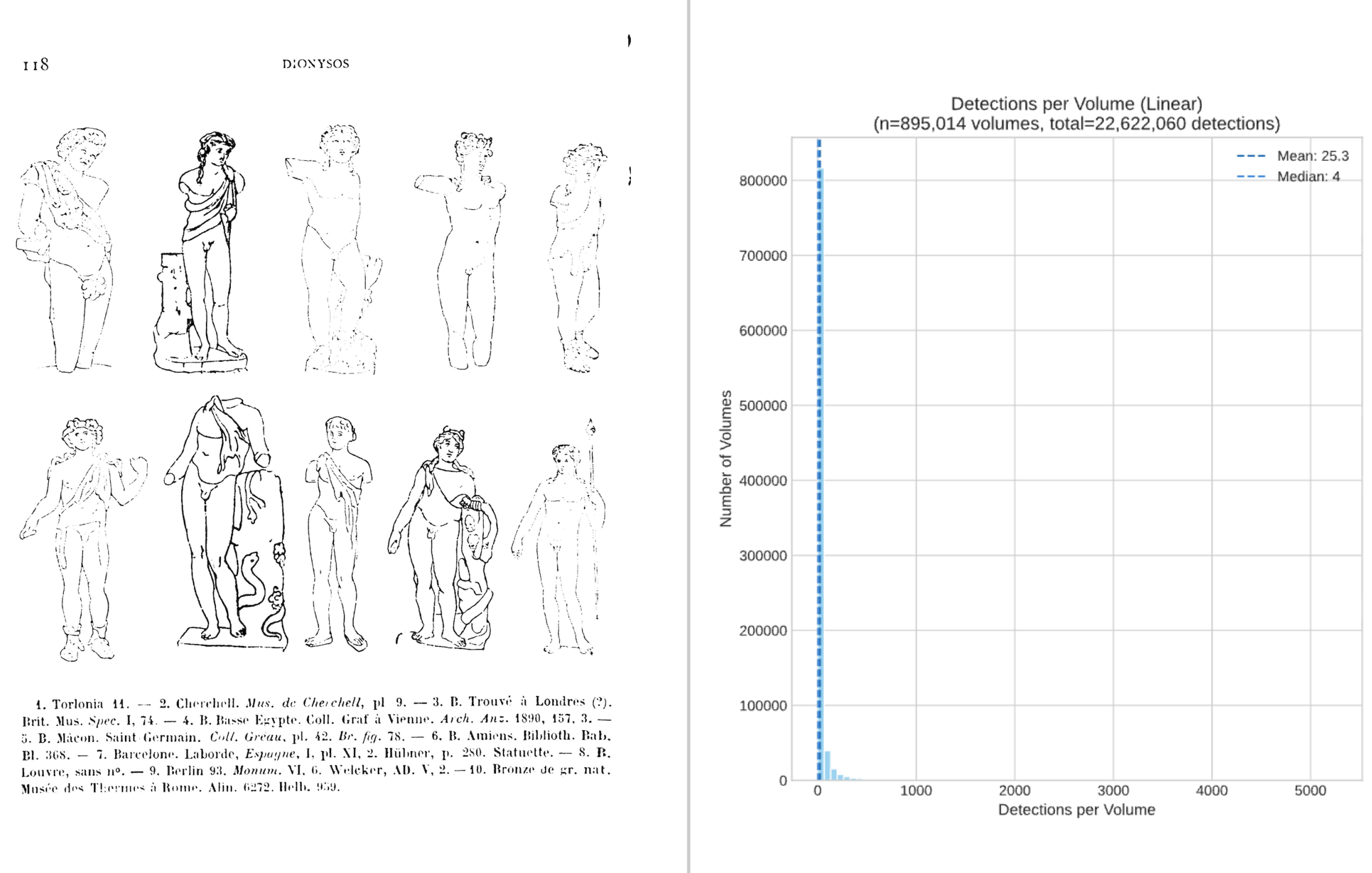

*Figure App. B1:* Screenshot.
Sample pages from volume in the collection with most detections. From barcode HXJU8K.

*Figure App. B2:* Linear-Scale Histogram.
Detections per volume. The linear scale highlights how most volumes had very few visual elements.

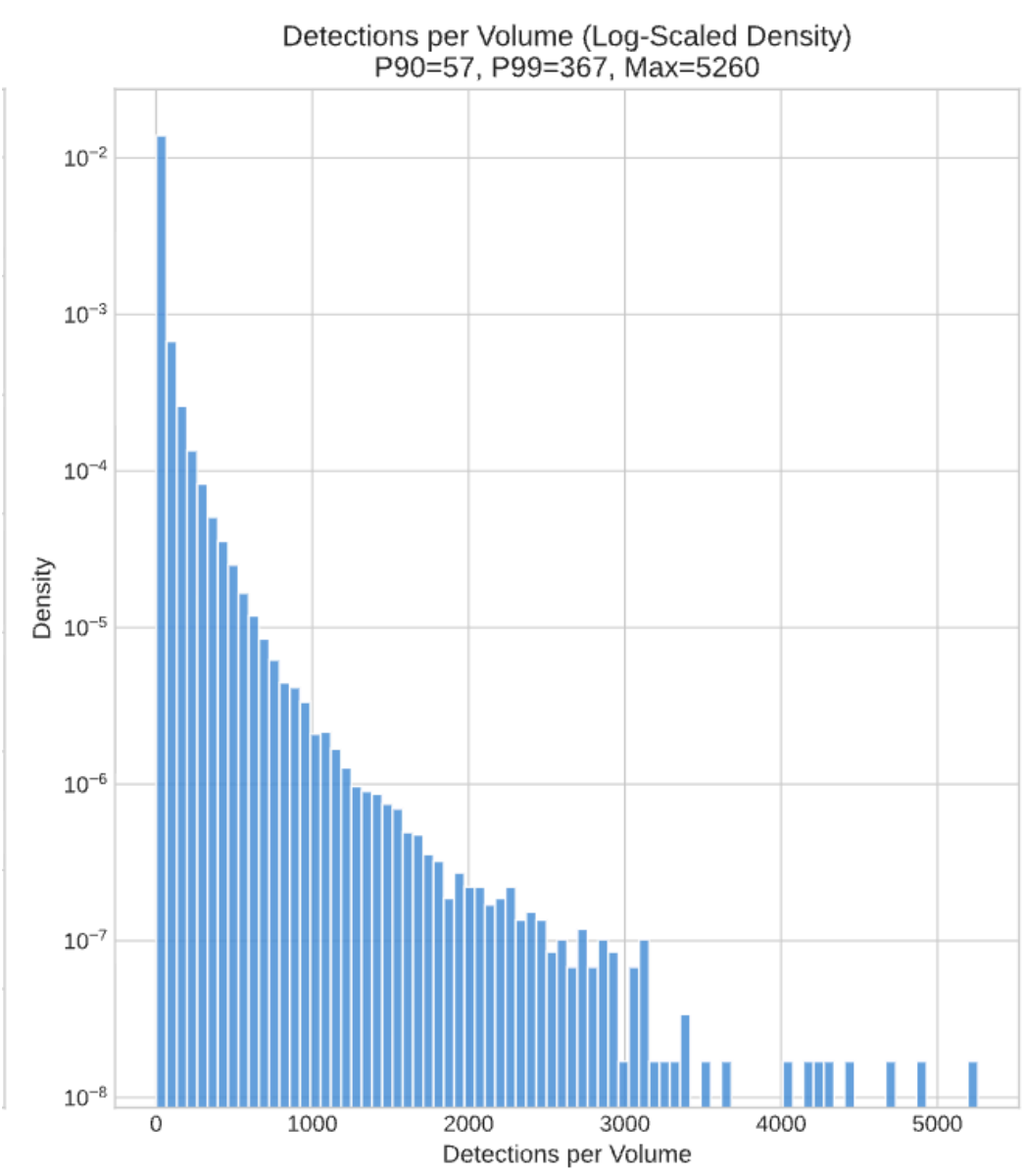

*Figure App. B3:* Log-Scale Histogram.
Detections per volume. The visual elements dataset is skewed toward a subset of volumes that contain a disproportionately large number of elements.

## Appendix C: Music reclassification keywords

*Table App. C: Keyword Lists for top 20 languages in "Music" class sorted by count*

| Language | Count | Keywords List |
|---|---|---|
| English | 758,663 | music, musical, sheet music, music notation, musical notation, musical notes, musical score, hymn, song, chorus, violin, piano, orchestra, choir |
| German | 165,748 | Musik, musikalisch, Noten, Musikalische Notation, Notenschrift, Noten, Partitur, Hymne, Lied, Chor, Geige, Klavier, Orchester, Chor |
| French | 154,740 | musique, musical, musicale, partition musicale, notation musicale, notation musicale, notes de musique, partition, hymne, chanson, refrain, violon, piano, orchestre, chorale |
| Italian | 28,005 | musica, musicale, partitura musicale, notazione musicale, notazione musicale, note musicali, spartito musicale, inno, canzone, coro, violino, pianoforte, orchestra, coro |
| Latin | 18,465 | musica, musicus, charta musica, scriptura musica, notatio musica, notae musicae, partitura musica, hymnus, carmen, chorus, violina, pianum, orchestra, chorus |
| Russian | 6,251 | музыка, музыкальный, ноты, музыкальная нотация, музыкальная запись, музыкальные ноты, партитура, гимн, песня, хор, скрипка, фортепиано, оркестр, хор |
| Swedish | 5,063 | musik, musikalisk, noter, notnotation, musiknotation, musiknoter, partitur, psalm, låt, kör, violin, piano, orkester, kör |
| Spanish | 4,817 | música, musical, partitura, notación musical, notación musical, notas musicales, partición musical, himno, canción, coro, violín, piano, orquesta, coro |
| Dutch | 4,809 | muziek, muzikaal, noten, muzieknotatie, muzikale notatie, muzikale noten, partituur, lied, nummer, refrein, viool, piano, orkest, koor |
| Polish | 4,411 | muzyka, muzyczny, nuty, zapis nutowy, notacja muzyczna, nuty muzyczne, partytura, hymn, piosenka, refren, skrzypce, pianino, orkiestra, chór |
| Czech | 4,080 | hudba, hudební, noty, hudební zápis, hudební notace, hudební noty, partitura, hymna, píseň, sbor, housle, piano, orchestr, sbor |
| Welsh | 4,016 | cerddoriaeth, cerddorol, nodi cerdd, notation cerdd, notation gerddorol, nodau cerdd, partitur cerdd, emyn, cân, corus, ffiḍil, piano, cherddllys, cor |
| Portuguese | 3,646 | música, musical, partitura, notação musical, notação musical, notas musicais, partitura musical, hino, canção, coro, violino, piano, orquestra, coral |
| Hebrew | 2,686 | מוזיקה, מוזיקלי, תווים, תיוון מוזיקלי, רישום מוזיקלי, תווים מוזיקליים, פרטיטורה, המנון, שיר, מקהלה, כינור, פסנתר, תזמורת, מקהלה |
| Hungarian | 2,474 | zene, zenés, kottalap, zenéírás, zenekotáció, hangjegyek, partitúra, himnusz, dal, karének, hegedű, zongora, zenekar, kórus |
| Danish | 1,726 | musik, musikalisk, nodebog, nodeskrift, musikalisk notation, noder, partitur, salme, sang, omkvæd, violin, klaver, orkester, kor |
| Armenian | 1,442 | երաժշտություն, երաժշտական, նոտադրություններ, երաժշտական նոտագրություն, ներդաշնակային նոտագրություն, ներդաշնակային նոտաներ, պարտիտուրա, հիմն, երգ, կորուս, ստրինգ, դաշնամուր, նվագախումբ, երգչախումբ |
| Croatian | 1,387 | glazba, glazbeni, notni zapis, izrada notacije, glazbena notacija, note, partitura, himna, pjesma, zbor, violina, klavir, orkestar, hor |
| Icelandic | 1,006 | tónlist, tónlistarlegur, tónblað, tónlistarskrif, tónlistarrit, tónar, tónverk, salmur, lag, kór, fiðla, flygil, hljómsveit, kór |
| Norwegian | 760 | musikk, musikalsk, noter, musikknotasjon, musikknottasjon, musikknoter, partitur, salme, sang, refreng, fiolin, piano, orkester, kor |

# Appendix D: Caption model parameters

*Table App. D: Caption model configuration*

| Parameter | Value | Rationale |
|---|---|---|
| Model | gpt-4.1-nano | Balance of capability and cost for high-volume processing |
| Temperature | 0 | Greedy decoding for consistency |
| Max Tokens | 100 | Enforces conciseness |
| Top Logprobs | 2 | Enables post-hoc analysis |
| Request Timeout | 20 seconds | Prevents indefinite hangs |
| Retry attempts | 2 | Handles transient API failures |

## Appendix E: Caption source and detected languages

*Table App. E1: Top 15 source languages*

| Language | ISO 639-3 | Count | Percent of total captions |
|---|---|---|---|
| English | eng | 11,163,719 | 60.35% |
| German | deu | 2,777,526 | 15.01% |
| French | fra | 2,314,030 | 12.51% |
| Italian | ita | 391,968 | 2.12% |
| Spanish | spa | 334,566 | 1.81% |
| Russian | rus | 134,541 | 0.73% |
| Swedish | swe | 134,024 | 0.72% |
| Portuguese | por | 91,082 | 0.49% |
| Dutch | nld | 80,781 | 0.44% |
| Czech | ces | 79,068 | 0.43% |
| Latin | lat | 76,068 | 0.41% |
| Danish | dan | 72,227 | 0.39% |
| Hungarian | hun | 57,219 | 0.31% |
| Chinese | zho | 51,394 | 0.28% |
| Polish | pol | 44,638 | 0.24% |

*Table App. E2: Top 15 detected languages*

| Language | ISO 639-3 | Count | Percent of total captions |
|---|---|---|---|
| English | eng | 11,200,292 | 60.55% |
| German | deu | 2,765,398 | 14.95% |
| French | fra | 2,304,942 | 12.46% |
| Italian | ita | 390,565 | 2.11% |
| Spanish | spa | 334,008 | 1.81% |
| Latin | lat | 135,920 | 0.73% |
| Swedish | swe | 134,652 | 0.73% |
| Russian | rus | 133,211 | 0.72% |
| Portuguese | por | 91,966 | 0.50% |
| Dutch | nld | 80,463 | 0.43% |
| Czech | ces | 79,177 | 0.43% |
| Danish | dan | 74,636 | 0.40% |
| Hungarian | hun | 57,465 | 0.31% |
| Chinese | zho | 51,458 | 0.28% |
| Polish | pol | 44,723 | 0.24% |

The slight mismatch between the detected language and source language represent failures of the model as well as misclassified languages.

## Appendix F: Orientation correction training report

*Table App. F: Training and validation performance per epoch*

| Epoch | Train Acc | Val Acc | Train Loss | Val Loss |
|---|---|---|---|---|
| 1 | 74.55% | 83.07% | 0.5919 | 0.4242 |
| 2 | 85.30% | 86.27% | 0.3597 | 0.3373 |
| 3 | 89.37% | 87.63% | 0.2642 | 0.3274 |
| 4 | 92.15% | 87.85% | 0.1985 | 0.3203 |
| 5 | 94.09% | 88.73% | 0.1527 | 0.3439 |
| 6 | 95.49% | 88.51% | 0.1174 | 0.3557 |
| 7 | 96.67% | 88.77% | 0.0891 | 0.3479 |
| 8 | 97.48% | 88.67% | 0.0707 | 0.3857 |
| 9 | 97.92% | 89.05% | 0.0570 | 0.3648 |
| 10 | 98.61% | 89.27% | 0.0422 | 0.4013 |
| 11 | 98.78% | 88.80% | 0.0342 | 0.4668 |
| 12 | 99.02% | 89.46% | 0.0283 | 0.4116 |
| 13 | 99.32% | 89.59% | 0.0197 | 0.4411 |
| 14 | 99.32% | 90.22% | 0.0202 | 0.4266 |
| 15 | 99.58% | 90.38% | 0.0128 | 0.4407 |
| 16 | 99.66% | 90.32% | 0.0094 | 0.4568 |
| 17 | 99.79% | 90.19% | 0.0074 | 0.4621 |
| 18 | 99.81% | 90.13% | 0.0062 | 0.4745 |
| 19 | 99.79% | 90.44% | 0.0060 | 0.4649 |
| 20 | 99.81% | 90.35% | 0.0058 | 0.4753 |

## Appendix G: Compute specs

*Table App. G: Compute resources*

| Component / Step | Hardware | Usage |
|---|---|---|
| Main visual-elements pipeline | Dual-socket AMD EPYC 9554 server with 755 GiB RAM and 4× NVIDIA L40S GPUs | 984 batches (up to 1,000 volumes each); 41 days 17 hours active pipeline time over 52 days total |
| Orientation correction | NVIDIA GH200 GPU | Inference on 16.7M eligible images over 2 days 23 hours. |
| Comparative smaller run of visual-elements pipeline | DGX-class system ("DGX Spark") | 9000 volumes processed end-to-end; ~23 hours 8 minutes total. |

## Appendix H: Temporal coverage breakdown

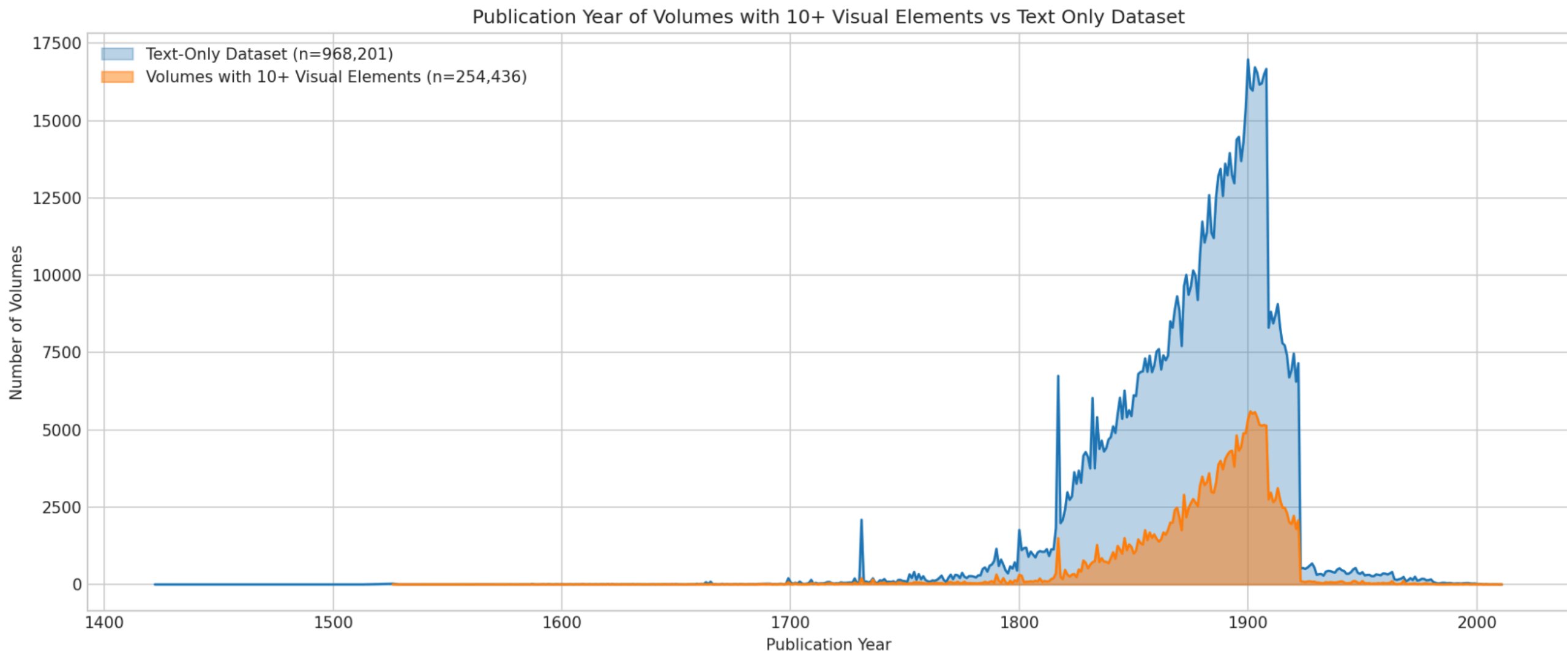

*Figure App. H:* Histogram
Approximate publication year distribution for volumes with at least 10 detections vs text-only dataset, using the Institutional Books: Harvard Library metadata as source of date (`date1_src` and `date2_src` as fallback). Only volumes with existing date values are used (254,436 volumes).

## Appendix I: Language coverage breakdown

*Table App. I: Top 20 languages in volumes with at least 10 detections compared to text-only dataset*

| Language | ISO 639-3 | Volume Count | Detection Count | % of Detections | % of Text-only Dataset |
|---|---|---|---|---|---|
| English | eng | 150,675 | 12,249,891 | 60.4% | 47.11% |
| German | deu | 35,671 | 3,145,802 | 15.5% | 14.66% |
| French | fra | 33,151 | 2,674,981 | 13.2% | 12.53% |
| Italian | ita | 8,047 | 483,795 | 2.4% | 4.36% |
| Spanish | spa | 5,747 | 423,679 | 2.1% | 2.73% |
| Russian | rus | 2,650 | 168,700 | 0.8% | 1.43% |
| Chinese | zho | 3,791 | 158,473 | 0.8% | 0.57% |
| Swedish | swe | 1,647 | 153,822 | 0.8% | 0.58% |
| Portuguese | por | 1,436 | 112,991 | 0.6% | 0.68% |
| Latin | lat | 1,962 | 102,903 | 0.5% | 1.96% |
| Dutch | nld | 1,845 | 96,994 | 0.5% | 1.20% |
| Czech | ces | 901 | 91,311 | 0.5% | 0.29% |
| Danish | dan | 1,138 | 80,622 | 0.4% | 0.52% |
| Hungarian | hun | 744 | 67,075 | 0.3% | 0.46% |
| Japanese | jpn | 825 | 65,690 | 0.3% | 0.17% |
| Polish | pol | 713 | 55,312 | 0.3% | 0.35% |
| Multiple languages | mul | 553 | 49,236 | 0.2% | 0.12% |
| Norwegian | nor | 533 | 39,124 | 0.2% | 0.22% |
| Undetermined | und | 670 | 38,901 | 0.2% | 0.34% |
| Greek | ell | 439 | 24,004 | 0.1% | 0.25% |

From volumes with at least 10 detections, using the Institutional Books 1.0 metadata as source of language (`language_src`) and text-only language percentage. Only volumes with non-empty language values are used (253,138 volumes).

## Appendix J: Topic coverage breakdown

*Table App. J: Count of topics for volumes with at least 10 detections*

| Topic / Subject | Number of Volumes | % in Visual Elements Dataset | % in Text Only Dataset |
|---|---|---|---|
| SCIENCE | 62,009 | 24.0% | 11.17% |
| LANGUAGE AND LITERATURE | 48,127 | 18.6% | 23.76% |
| MEDICINE | 17,752 | 6.9% | 3.21% |
| PHILOSOPHY. PSYCHOLOGY. RELIGION | 16,738 | 6.5% | 11.58% |
| AGRICULTURE | 14,048 | 5.4% | 3.70% |
| AUXILIARY SCIENCES OF HISTORY | 12,242 | 4.7% | 3.42% |
| FINE ARTS | 12,077 | 4.7% | 2.23% |
| LAW | 10,838 | 4.2% | 12.94% |
| TECHNOLOGY | 10,306 | 4.0% | 1.69% |
| GEOGRAPHY. ANTHROPOLOGY. RECREATION | 9,519 | 3.7% | 2.36% |
| HISTORY OF THE AMERICAS | 9,110 | 3.5% | 2.73% |
| SOCIAL SCIENCES | 7,771 | 3.0% | 5.10% |
| MUSIC AND BOOKS ON MUSIC | 6,876 | 2.7% | 1.32% |
| EDUCATION | 6,161 | 2.4% | 2.29% |
| POLITICAL SCIENCE | 3,543 | 1.4% | 2.72% |
| GENERAL WORKS | 3,039 | 1.2% | 0.67% |
| MILITARY SCIENCE | 2,514 | 1.0% | 0.69% |
| NAVAL SCIENCE | 2,174 | 0.8% | 0.51% |
| WORLD HISTORY AND HISTORY OF EUROPE, ASIA, | 1,716 | 0.7% | 0.73% |
| AFRICA, AUSTRALIA, NEW ZEALAND, ETC. | | | |
| BIBLIOGRAPHY. LIBRARY SCIENCE. INFORMATION RESOURCES (GENERAL) | 1,672 | 0.6% | 0.55% |

From volumes with at least 10 detections, using the Institutional Books metadata as source of topic information (`topic_or_subject_gen`). Only volumes with non-empty topic values are used (258,232 volumes).